\documentclass[journal]{IEEEtai}

\usepackage[colorlinks,urlcolor=blue,linkcolor=blue,citecolor=blue]{hyperref}
\usepackage{amsmath,amsfonts}
\usepackage{algorithmic}
\usepackage{algorithm}
\usepackage{array}
\usepackage[caption=false,font=normalsize,labelfont=sf,textfont=sf]{subfig}
\usepackage{textcomp}
\usepackage{stfloats}
\usepackage{url}
\usepackage{verbatim}
\usepackage{graphicx}
\usepackage{cite}
\usepackage{multirow}
\usepackage{hyperref}
\hypersetup{
    colorlinks=true,
    linkcolor=blue,
    filecolor=blue,      
    urlcolor=blue,
    citecolor=cyan,
}
\newcommand{\eg}{\textit{e}.\textit{g}.}
\newcommand{\etal}{\textit{et al}.}
\newcommand{\ie}{\textit{i}.\textit{e}.}

\usepackage{color,array}

\begin{document}

\title{360$^\circ$ High-Resolution Depth Estimation via Uncertainty-aware Structural Knowledge Transfer}

\author{Zidong Cao, Hao Ai, Athanasios V. Vasilakos,~\IEEEmembership{IEEE Senior Member},  Lin Wang$^\dagger$,~\IEEEmembership{Member,~IEEE}
        % <-this % stops a space
\thanks{Z. Cao, H. Ai are with the Artificial Intelligence Thrust, The Hong Kong University of Science and Technology (HKUST), Guangzhou, China. E-mail: \{caozidong1996@gmail.com, hai033@connect.hkust-gz.edu.cn\}}
\thanks{Athanasios V. Vasilakos is with the Center for AI Research (CAIR), University of Agder(UiA), Grimstad, Norway. Email: thanos.vasilakos@uia.no}
\thanks{L. Wang is with the Artificial Intelligence Thrust, HKUST, Guangzhou, and Dept. of Computer Science and Engineering, HKUST, Hong Kong SAR, China. E-mail: linwang@ust.hk.}

\thanks{$^\dagger$Corresponding author}}

\markboth{Journal of IEEE Transactions on Artificial Intelligence, Vol. 00, No. 0, Month 2020}
{Z. Cao. \MakeLowercase{\textit{et al.}}: 360$^\circ$ High-Resolution Depth Estimation via Uncertainty-aware Structural Knowledge Transfer}

\maketitle

\begin{abstract} 
To predict high-resolution (HR) omnidirectional depth maps, existing methods typically leverage HR omnidirectional image (ODI) as the input via fully-supervised learning. However, in practice, taking HR ODI as input is undesired due to resource-constrained devices. In addition, depth maps are often with lower resolution than color images.
Therefore, in this paper, we explore for the first time to estimate the HR omnidirectional depth directly from a low-resolution (LR) ODI, when no HR depth GT map is available. 
Our key idea is to transfer the scene structural knowledge from the HR image modality and the corresponding LR depth maps to achieve the goal of HR depth estimation without any extra inference cost.
Specifically, we introduce ODI super-resolution (SR) as an auxiliary task and train both tasks collaboratively in a weakly supervised manner to boost the performance of HR depth estimation. 
The ODI SR task extracts the scene structural knowledge via uncertainty estimation. 
Buttressed by this, a scene structural knowledge transfer (SSKT) module is proposed with two key components.
First, we employ a cylindrical implicit interpolation function (CIIF) to learn cylindrical neural interpolation weights for feature up-sampling and share the parameters of CIIFs between the two tasks. 
Then, we propose a feature distillation (FD) loss that provides extra structural regularization to help the HR depth estimation task learn more scene structural knowledge.
Extensive experiments demonstrate that our weakly-supervised method outperforms baseline methods, and even achieves comparable performance with the fully-supervised methods. 
\end{abstract}

\begin{IEEEImpStatement} 
Our weakly-supervised can adapt to various backbones, input sizes, and datasets. Our method can also easily be applied to higher resolution, such as $1024\times2048$. We contribute the effectiveness to the proposed SSKT module, which successfully transfers essential scene structural knowledge from the ODI SR task to the HR depth estimation task. Importantly, we prove that the way of transferring the scene structural knowledge from the HR image modality and corresponding LR depth map is feasible to achieve the goal of HR depth estimation without any extra inference cost.
\end{IEEEImpStatement}


\section{Introduction}

% \begin{figure}[htbp]
%     \centering
%     \includegraphics[width=0.94\linewidth]{Figure/intro.pdf}
%     \caption{\textbf{Visual comparison of omnidirectional monocular depth estimation results with $\times8$ up-sampling factor on Stanford2D3D dataset}~\cite{Armeni2017Joint2D}. Left: predicted depth maps, Right: reconstructed point clouds. (a) Our method without uncertainty estimation. (b) Our method without FD loss. (c) Our method (All). (d) UniFuse-Fusion~\cite{Jiang2021UniFuseUF} with fully supervision. (e) Ground truth.}
%     \label{fig:intro}
% \end{figure}

\begin{figure}[t!]
    \centering
    \includegraphics[width=0.85\linewidth]{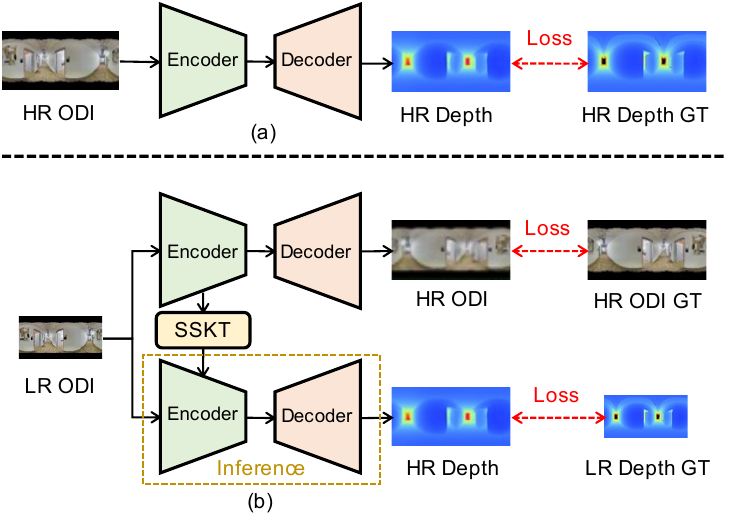}
    \vspace{-10pt}
    \caption{HR depth estimation paradigms. (a) Existing fully-supervised methods. (b) Our weakly-supervised method transfers knowledge between two tasks via the SSKT module.}
    \vspace{-15pt}
    \label{fig:comp}
\end{figure}

Monocular depth estimation is a fundamental task of 3D scene understanding that finds its applications in various fields, such as autonomous driving. The advancements in deep neural networks (DNNs)~\cite{galvan2021neuroevolution} and the accessibility of large-scale annotated datasets have resulted in notable improvements in monocular depth estimation. Nevertheless, most of the existing approaches are developed for perspective images, which suffer from a limited field-of-view (FoV)~\cite{Eigen2014DepthMP,ai2022deep}. 

Recently, $360^{\circ}$ cameras\footnote{Also called omnidirectional cameras.} have received considerable attention because of their ability to capture the environment with a wide FoV of $180^\circ \times 360^\circ$ in a single shot. Omnidirectional images (ODIs)\footnote{ODIs usually indicate the equirectangular projection (ERP) type images.} present the scene entirely with rich scene contexts. This distinct advantage has inspired active research for monocular depth estimation~\cite{sun2021hohonet,Jiang2021UniFuseUF,Ai2023HRDFuseM3,Wang2020BiFuseM3}.
% tasks such as room layout estimation~\cite{zeng2020joint, wang2021led2}, scene understanding~\cite{sun2021hohonet, jin2020geometric}, simultaneous localization and mapping (SLAM)~\cite{won2020omnislam}, and virtual reality (VR). 
However,
% , the wide FoV can be a two-edged sword.
the angular resolution of ODIs is relatively lower than that of perspective images when ODIs and perspective images are captured by different sensors with identical sizes. The low angular resolution leads to degraded structural details such as edges, rendering it difficult for omnidirectional depth estimation.
% While there are state-of-the-art omnidirectional monocular depth estimation approaches~\cite{Jiang2021UniFuseUF, Wang2020BiFuseM3,Zioulis2018OmniDepthDD} that can predict omnidirectional depth maps, they often do so in limited resolutions, such as $512 \times 1024$. Although such resolution may suffice for certain tasks like room layout estimation, it is inadequate to provide an acceptable quality of immersive experience for VR applications. Generally, VR applications require at least the resolution of $10800 \times 21600$ to match the existing head mounted displays (HMDs)~\cite{ozcinar2019super}.

To reconstruct the structural details, existing methods~\cite{Wang2020BiFuseM3, Jiang2021UniFuseUF,li2022omnifusion,Ai2023HRDFuseM3,Zioulis2018OmniDepthDD,ReyArea2021360MonoDepthH3,Tateno2018DistortionAwareCF,zhuang2021acdnet,peng2023high}
% are typically based on an encoder-decoder architecture to ensure that the predicted depth map has consistent resolution with the input ODI. 
leverage high-resolution (HR) ODIs as inputs with the supervision of HR depth ground truth (GT) maps. For instance, using the HR ODIs and paired HR depth GT maps, OmniDepth~\cite{Zioulis2018OmniDepthDD} employs the encoder-decoder architecture to predict HR depth maps, as shown in Fig.~\ref{fig:comp}(a).
% to combine the features extracted from different projection formats, \ie, equirectangular projection (ERP) and cubemap projection, of HR ODIs to predict HR
However, in practice, taking HR ODI as input is undesired by resource-constrained devices. 
% Specifically, to provide immersive experience in applications such as virtual reality (VR), ODIs are required with higher resolution than 2D counterparts. 
Take virtual reality (VR) as an example, the headsets often have limited bandwidth for transmission and limited computational resources~\cite{mangiante2017vr}. Processing HR ODIs in the headsets might cause high latency due to limited bandwidth~\cite{mangiante2017vr}. In addition,
obtaining HR depth maps is arduous and expensive~\cite{Sun2021LearningSS}. This is because depth sensors, such as LiDARs, often capture sparse depths with low resolution~\cite{geiger2012we}. Also, increasing the lens of LiDARs is far more expensive than obtaining HR ODI~\cite{li2020lidar}.

 \textbf{Motivation.} In this paper, \textit{we explore for the first time to estimate the HR depth map directly from an LR ODI, when no HR depth GT map is available} (See Fig~\ref{fig:comp}(b)). Without the supervision of HR depth GT maps, the fine scene structural details can scarcely be recovered in the HR depth predictions. 
% The main challenge is that without HR depth GT map, fine scene structural details are usually lost or destroyed in the HR depth estimation~\cite{Sun2021LearningSS}.
Therefore, to explore valuable supplementary supervision that can facilitate HR depth estimation, \textit{our key idea is to learn and transfer scene structural knowledge from the HR image modality and the corresponding LR depth maps to achieve the goal of HR depth estimation without causing extra inference cost}. This is inspired by the fact that \textbf{1)}
% HR ODIs are easier to be captured, 
% than HR omnidirectional depth maps, 
there exists publicly available large-scale LR-HR paired image datasets and thus achieve the super-resolution (SR) task~\cite{Deng2021LAUNetLA, Yoon2021SphereSR3I,zhang2022heat, zhang2022curvature, zhang2022fluid}; \textbf{2)} an ODI and its corresponding depth map are photometric and geometric representations of the same scene, respectively~\cite{Sun2021LearningSS}. This implies that both representations share the common scene structural knowledge, such as edges.

% \begin{figure}[t!]
%     \centering
%     \includegraphics[width=0.8\linewidth]{Figure/comp.pdf}
%     \caption{HR depth estimation paradigms. (a) Existing fully-supervised methods. (b) Our weakly-supervised method transfers knowledge between two tasks via the SSKT module.}
%     \vspace{-15pt}
%     \label{fig:comp}
% \end{figure}

% \begin{figure}[t]
%     \centering
%     \includegraphics[width=0.94\linewidth]{Figure/intro.pdf}
%     \caption{\textbf{Visual comparison of omnidirectional monocular depth estimation results with $\times8$ up-sampling factor on Stanford2D3D dataset}~\cite{Armeni2017Joint2D}. Left: predicted depth maps, Right: reconstructed point clouds. (a) Our method without uncertainty estimation. (b) Our method without FD loss. (c) Our method (All). (d) UniFuse-Fusion~\cite{Jiang2021UniFuseUF} with fully supervision. (e) Ground truth.}
%     \label{fig:intro}
% \end{figure}

Specifically, we employ the ODI SR as an auxiliary task and train both tasks collaboratively in a weakly-supervised manner to boost the performance of HR depth estimation. The ODI SR task takes an LR ODI as input to predict a super-resolved RGB image. As the textural details in the super-resolved RGB image may interfere with the smoothness of depth prediction~\cite{karsch2014depth, Jiang2021UniFuseUF}, we introduce uncertainty estimation~\cite{ning2021uncertainty} for the ODI SR task to specially extract the scene structural knowledge. The high-uncertainty regions are often matched with high-frequency regions, which can be utilized to emphasize the regions carrying more structural knowledge~\cite{ning2021uncertainty}. Upon this, we propose a scene structural knowledge transfer (SSKT) module with two key components. 

Firstly, considering the unique properties of the equirectangular projection (ERP) type of ODIs, we design a novel cylindrical implicit interpolation function (CIIF) for feature up-sampling by learning the neural interpolation weights based on cylindrical coordinates. This way, our CIIF alleviates the discontinuity and stretched distortion of ODIs. Also, as the learned interpolation weights can capture the correlations among neighboring pixels, they can be used to represent the scene structural knowledge.
% , which closely relate to scene structural knowledge. 
Therefore, by sharing the parameters of CIIFs between the ODI SR and HR depth estimation tasks, structural knowledge can be transferred between the two tasks effectively.
Secondly, as the features of the ODI SR task also contain crucial scene structural knowledge, we propose a feature distillation (FD) loss to refine depth estimation.

In total, our weakly-supervised method is \textit{efficient}, as the ODI SR network and SSKT module can be removed freely and only LR ODI is needed during inference, introducing no extra inference cost. Moreover, our method is \textit{flexible} and can be easily adapted as a plug-and-play approach by modifying the backbones for the HR depth estimation network and varying the up-sampling factors (See Tab.~\ref{table:comparison}). Furthermore, we demonstrate the \textit{effectiveness} of our method across different datasets~\cite{Armeni2017Joint2D,chang2017matterport3d,Zioulis2018OmniDepthDD}, with various up-sampling factors and backbones. 
The experimental results show that our method not only \textit{outperforms} the baseline methods, but also achieves \textit{comparable} performance with the fully-supervised methods. Our main contributions can be summarized into four-fold:

\begin{itemize}
\item We propose the first weakly-supervised method that predicts the HR omnidirectional depth map from an LR ODI when no HR depth GT map is available.
\item We propose to learn and transfer the scene structural knowledge from an uncertainty-driven ODI SR task.
\item We propose an SSKT module with two components. The first is the CIIF as it is specific for ODIs to learn neural interpolation weights. The second is the FD loss to provide extra structural regularization.
\item Experimental results demonstrate that our proposed method is efficient, flexible, and effective, which achieves comparable results with the fully supervised methods.
\end{itemize}

\begin{figure*}[t!]
    \centering
    \includegraphics[width=0.85\textwidth]{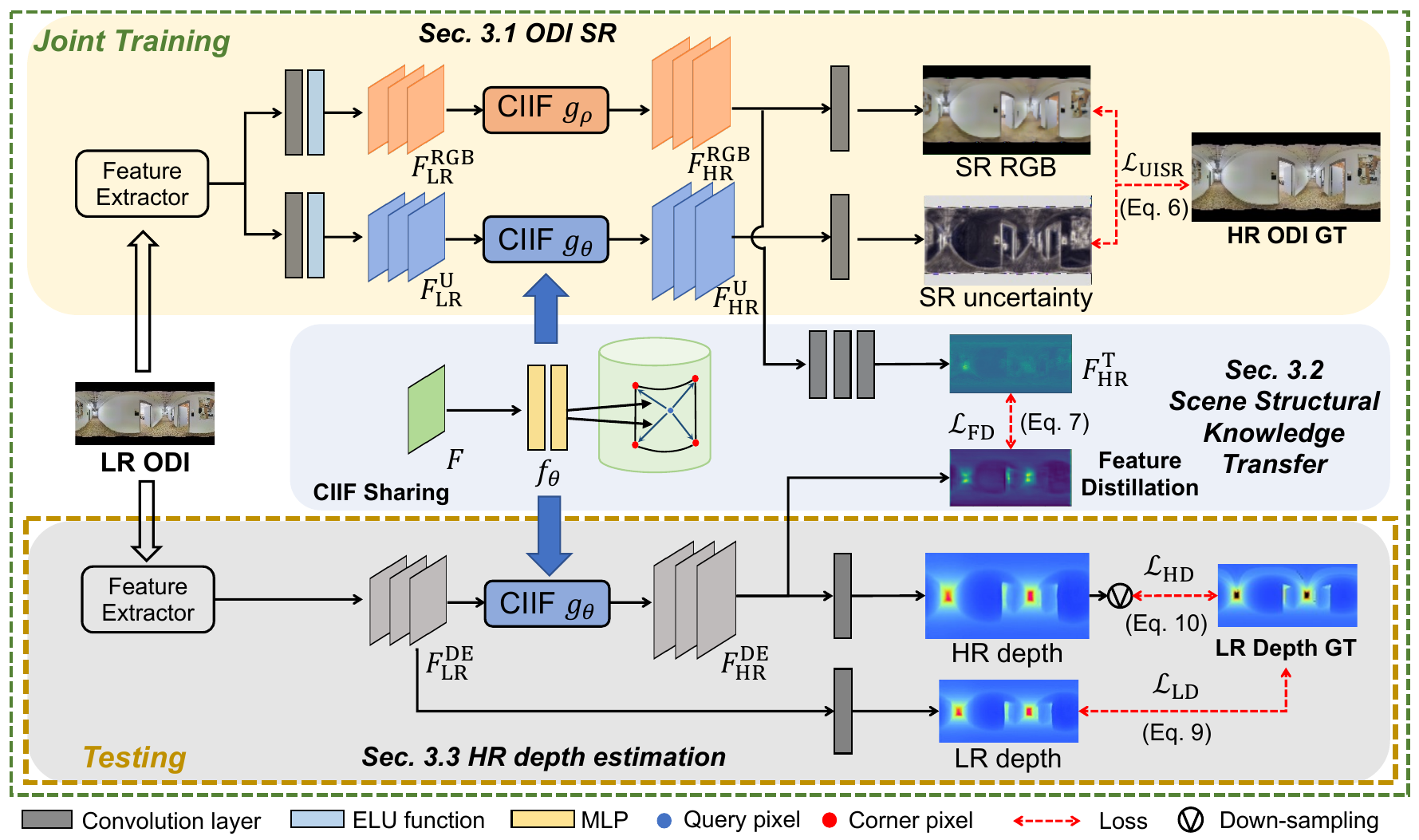}
    \vspace{-5pt}
    \caption{\textbf{Overview of our weakly-supervised framework}. Firstly, we introduce an ODI SR task (Sec~\ref{ISR-UL}) which predicts uncertainty to extract structural knowledge. Then, we design an SSKT module with CIIF that learns cylindrical neural interpolation weights and shares parameters between the two tasks. The SSKT module also includes an FD loss (Sec~\ref{CMKT}) for feature distillation. Finally, we employ an HR depth estimation task (Sec~\ref{DESR}) to generate HR depth estimation directly from an LR ODI.
     The detailed components of CIIF is shown in Fig.~\ref{fig:CIIF}(b).}
    \label{fig:framework}
    \vspace{-15pt}
\end{figure*}

\section{Related Work}
% \subsection{Omnidirectional Monocular Depth Estimation}
\noindent \textbf{Omnidirectional Monocular Depth Estimation.}
% Omnidirectional monocular depth estimation has evolved rapidly, as it owns the high potential for 3D vision tasks. 
Some attempts have been made to tackle the spherical distortion in ODIs by modifying the characteristics of convolutional filters, including row-wise rectangular filter~\cite{Zioulis2018OmniDepthDD}, deforming the sampling grids~\cite{Tateno2018DistortionAwareCF}, and dilated convolutions for larger receptive fields~\cite{zhuang2021acdnet}. 
Another direction is to combine different projection types, \eg, ERP, cube map~\cite{Wang2020BiFuseM3,Jiang2021UniFuseUF}, and tangent projection~\cite{li2022omnifusion,shen2022panoformer}.
BiFuse~\cite{Wang2020BiFuseM3} proposes a bi-projection fusion model for the ERP images and cube maps to share information with each other, while UniFuse~\cite{Jiang2021UniFuseUF} demonstrates that the unidirectional fusion from the cube map to ERP is more effective. Some methods~\cite{ReyArea2021360MonoDepthH3,peng2023high} focus on parallel computation in
% high-resolution
omnidirectional monocular depth estimation. The typical pipeline is first projecting an ERP image into multiple low-resolution perspective patches, then predicting depth maps in parallel, and finally re-projecting the perspective depth predictions back to the ERP plane. Recently, Li \etal~\cite{li2023mathcal} propose to extract features on the unit sphere to relieve the distortion effect. However, all the aforementioned methods require HR depth maps for supervision. 
\noindent\textbf{Omnidirectional image super-resolution.}
% Early ODI SR methods \cite{Arican2011JointRA,Bagnato2010PlenopticBS,Nagahara2000SuperresolutionFA} assemble multiple LR ODIs to reconstruct an HR one. However, they rely heavily on the precision of calibration and the number of inputs. 
Cagri \etal~\cite{Ozcinar2019SuperresolutionOO} propose the first DNN-based framework via adversarial learning. LAU-Net~\cite{Deng2021LAUNetLA} addresses the non-uniformly distributed pixel density of the ERP type ODI via adaptively up-sampling different latitude bands. SphereSR~\cite{Yoon2021SphereSR3I} proposes a continuous spherical representation to model the sphere as an icosahedron and reconstruct ODIs with multiple projection types. 
% Recently, OSRT~\cite{yu2023osrt} and OPDN~\cite{sun2023opdn} employ distortion-aware transformers to alleviate the effect of distortion. 

\noindent\textbf{Modeling Uncertainty in image super-resolution.} 
In the image SR task, high-frequency regions with rich texture and edges carry more visual information than the smooth regions~\cite{ning2021uncertainty}. To better capture the high-frequency details and emphasize the corresponding pixels, some uncertainty-driven image SR methods have been proposed~\cite{Chung2019GRAMGR,ning2021uncertainty}. For instance, GRAM~\cite{Chung2019GRAMGR} extracts the uncertainty map via the Gaussian assumption and takes the uncertainty map as the attenuation mask to achieve better performance. Differently, Ning \etal~\cite{ning2021uncertainty} estimate the uncertainty with Jeffrey’s prior and proposes an uncertainty-driven loss to provide attention for the pixels with high uncertainty. Our method introduces the uncertainty estimation to extract scene structural knowledge, which is transferred to the depth estimation task.
% High-uncertainty regions are often matched with the high-frequency regions, and are commonly utilized to emphasize the corresponding pixels.
% In general, uncertainty is often assumed to be compatible with the Laplace or Gaussian distributions, and utilized to prioritize the high-frequency regions in the loss functions. 
% Specifically, by assuming that uncertainty is compatible with the Laplace or Gaussian distributions, GRAM~\cite{Chung2019GRAMGR} takes uncertainty as the attention mask in the loss, while~\cite{ning2021uncertainty} inverses the pre-trained uncertainty as the loss weights. Differently, considering that textural details can interfere the smoothness of depth maps, we estimate uncertainty to separate high-frequency structural details from high-frequency textural details.  

% \subsection{Implicit Neural Representation} 
\noindent \textbf{Implicit Neural Representation.} For the image SR task, Chen \etal~\cite{chen2021learning} propose an implicit function, LIIF, to represent images continuously and achieves the up-sampling process with arbitrary factors. Furthermore, Tang \etal~\cite{tang2021joint} replace the non-parametric interpolation weights in LIIF with the ones learned from the input image and depth map to achieve the guided depth SR task. However, the mentioned two implicit functions~\cite{chen2021learning, tang2021joint} are based on the Cartesian coordinate system, which is weak in tackling the discontinuity and distortion problems in the ERP type ODIs. In contrast, our proposed CIIF is constructed based on cylindrical coordinates, which is effective in alleviating the mentioned problems in ERP type ODIs, as evaluated in Tab.~\ref{table:ablation-interpolation}. 

% Importantly, CIIF focuses on predicting neural interpolation weights, which enables the parameters of CIIF to be shared among different modalities with geometric similarities.

% which are related to structural knowledge consistent in different modalities and the parameters of CIIF is possible to be shared among different modalities.

% utilizes the implicit representation for guided depth super-resolution. It outputs depth values and refined interpolation weights. However, it requires HR image as input and HR depth ground-truth for supervision, which is not applicable in our task. Recently,~\cite{lee2022local,lee2022learning} extend the implicit representation to the Fourier space to enhance the high-frequency representation. In contrast, our CIIF is constructed based on cylindrical coordinate to address the discontinuity and distortion problems in ERP images. The prediction of CIIF is not RGB value or depth value, but the interpolation weights, which relate to structural knowledge consistent in different modalities. Therefore, the parameters of CIIF is possible to be shared among different modalities.

% \subsection{Multi-task Learning}
% \noindent\textbf{Multi-task Learning.}
\noindent \textbf{Multi-task learning.} It is mainly used for correlated tasks~\cite{niu2020decade} or domains~\cite{zhang2018dual, zhang2019deep, zhang2019neural}. 
% Liu \etal~\cite{Liu20213Dto2DDF} leverage 3D features to enhance 2D features extracted from the images. 
Wang \etal~\cite{Wang2020DualSL} treat image SR as an auxiliary task for semantic segmentation with full supervision. Sun \etal~\cite{Sun2021LearningSS} leverage the monocular depth estimation task to facilitate the depth SR task, reconstructing finer details. 

\section{Method}
\noindent \textbf{Overview.}
The goal is to predict an HR omnidirectional depth map $D_{\mathrm{HR}}$ from an LR ODI $I_{\mathrm{LR}}$ as input, using only an LR depth map $D^{\mathrm{GT}}_{\mathrm{LR}}$ for supervision. As shown in Fig.~\ref{fig:framework}, our framework consists of an ODI SR task, a scene structural knowledge transfer (SSKT) module, and an HR depth estimation task. The ODI SR is employed as an auxiliary task and trained collaboratively with the HR depth estimation task. Specifically, we introduce uncertainty estimation to the ODI SR task for scene structural knowledge extraction. To transfer the extracted knowledge, the SSKT module is proposed with two components. The first is to share the parameters of the proposed cylindrical implicit interpolation function (CIIF) between the two tasks. The second is the feature distillation (FD) loss that provides extra structural regularization. The overall training process can be seen in Algorithm.~\ref{alg}. We now describe the components in detail.

% Specifically, in the UISR branch, we estimate uncertainty to separate structural details from textural details. In the feature up-sampling process, we propose the cylindrical implicit interpolation functions (CIIFs) to learn structural-aware interpolation weights based on cylindrical coordinate. The image and uncertainty features are up-sampled through CIIFs with the parameters $g_{\beta}$ and $g_{\theta}$, respectively.  Then, in the CMKT module, we share the parameters $g_{\theta}$ of CIIF for the uncertainty feature up-sampling with that for depth feature up-sampling, which is a hard method of knowledge transfer. Moreover, we propose the feature distillation (FD) loss (Eq.~\ref{eq14}) to strengthen the correlation between the HR image feature and HR depth feature as a soft method. Finally, with the transferred knowledge from the UISR branch and CMKT module, the OMDE branch generates HR depth prediction $D_{\mathrm{HR}}$ with weakly-supervised loss functions Eq.~\ref{eq-loss_lrde},\ref{eq-loss_srde}. We now describe these components in detail. 

\subsection{ODI SR}
\label{ISR-UL}

The omnidirectional images contain not only the complete structural details of the surroundings but also rich textural details in the scenes. However, the enriched textural details can interfere with the smoothness of predicted depth maps~\cite{Ai2023HRDFuseM3, Jiang2021UniFuseUF,karsch2014depth}. Therefore, directly transferring knowledge from the ODI SR task degrades the depth estimation performance~(See Tab.~\ref{table:ablation-share}).
Inspired by the recent progress of uncertainty-driven image super-resolution~\cite{ning2021uncertainty}, which demonstrates that regions with high uncertainty are often matched with high-frequency regions, we introduce the uncertainty estimation to the ODI SR task. By doing so, the regions that carry more structural knowledge can be emphasized, which is beneficial to reconstruct more structural details in the HR depth estimation task. Specifically, as shown in Fig.~\ref{fig:framework}, we utilize the feature extractor to process the LR ODI $I_{\mathrm{LR}}$ and utilize two groups of convolutional layers with ELU activation functions~\cite{Clevert2015FastAA} to output LR features for the RGB part $F_{\mathrm{LR}}^{\mathrm{RGB}}$ and uncertainty part $F_{\mathrm{LR}}^{\mathrm{U}}$, respectively. Then, $F_{\mathrm{LR}}^{\mathrm{RGB}}$ is up-sampled by CIIF with parameters $g_{\rho}$ to generate HR features $F_{\mathrm{HR}}^{\mathrm{RGB}}$.
 Similarly, $F_{\mathrm{LR}}^{\mathrm{U}}$ is up-sampled by CIIF with parameters $g_{\theta}$ to generate HR features $F_{\mathrm{HR}}^{\mathrm{U}}$.
We then describe the components of CIIF that achieve the feature up-sampling process.

\begin{figure}[t!]
    \centering
    \includegraphics[width=0.85\linewidth]{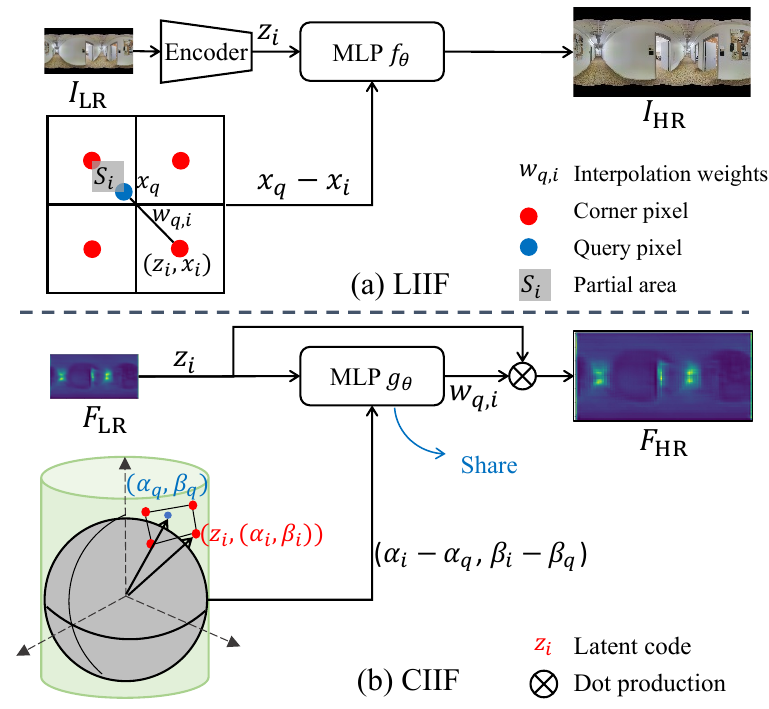}
    \caption{(a) LIIF predicts the RGB value based on the Cartesian coordinates. (b) Our CIIF learns neural interpolation weights based on the cylindrical coordinates. The parameters of CIIF $g_{\theta}$ are contained in the MLP shared from the ODI SR task.}
    \label{fig:CIIF}
    \vspace{-20pt}
\end{figure}

\noindent\textbf{CIIF.} 
 Generally, the image interpolation problems can be defined as follows:
 \setlength{\abovedisplayskip}{3pt}
\setlength{\belowdisplayskip}{3pt}
\begin{equation}
I_{\mathrm{HR}}(x_q) = \sum_{i \in N_q}{w_{q,i}I_{\mathrm{LR}}(x_i)},
\label{eq-interpolation}
\end{equation}
where $x_q$ is the coordinate of query pixel $q$ in the HR image domain, $x_i$ is the coordinate of corner pixel $i$ in the LR image domain; $N_q$ represents the neighboring pixels of $q$ in the LR image domain and $w_{q,i}$ is the interpolation weight between pixel $q$ and $i$. We take LIIF~\cite{chen2021learning} as an example to illustrate the interpolation process, depicted in Fig.~\ref{fig:CIIF}(a). In LIIF, given the latent code $z_i$ encoded from $I_{\mathrm{LR}}$ and the coordinate relation between $q$ and $i$, the RGB value of  $q$ in the HR image domain is predicted by the MLP $f_{\theta}$ with parameters $\theta$ as follows:
\setlength{\abovedisplayskip}{3pt}
\setlength{\belowdisplayskip}{3pt}
\begin{equation}
I_{\mathrm{HR}}(x_q) = f_{\theta}(z_i,x_q-x_i).
\label{eq-liif}
\end{equation}

To address the border discontinuity issue, LIIF~\cite{chen2021learning} averages the four predictions of query pixel $q$ from its four corner pixels. Notably, the interpolation weight $w_{q,i}$ is calculated by the partial area $S_i$ diagonally opposite to the corner pixel $i$ (See Fig.~\ref{fig:CIIF}(a)). Thus, Eq.~\eqref{eq-liif} can be extended to:
\begin{equation}
I_{\mathrm{HR}}(x_q) = \sum_{i=1}^{4} \frac{S_i}{\sum{S_i}}f_{\theta}(z_i, x_q-x_i).
\label{eq-area}
\end{equation}

Furthermore,~\cite{tang2021joint} replaces the scale factor $S_i$ in Eq.~\eqref{eq-area} with learnable interpolation weights to obtain smoother results.
However, for ODIs with ERP type, which unfolds from the sphere in a cylindrical manner, there are two specific challenges: 1) The discontinuity between the left and right sides; 2) The stretched distortion toward the poles (the top and bottom of ODIs). Previous implicit interpolation methods, \eg~\cite{chen2021learning,tang2021joint}, are based on the Cartesian coordinate, which is less effective in representing spatial correlation of pixels in ODIs~\cite{Yoon2021SphereSR3I}.
% aforementioned methods are limited to process the ERP images as they are based on the Cartesian coordinate and powerless to represent spatial correlation of pixels in the ERP images~\cite{Yoon2021SphereSR3I}.
Therefore, we propose to reframe ODIs with the cylindrical representation, as shown in Fig.~\ref{fig:CIIF}(b). The cylindrical angle coordinate is formulated as $\alpha\in(-\arctan(\frac{\pi h}{w}), \ \arctan(\frac{\pi h}{w}))$ (Latitude angle), $\beta\in(-\pi,\ \pi)$ (Longitude angle). For the calculation of the cylindrical representation, we recommend readers to refer to the first section of the supplemental material.
Based on the cylindrical angle coordinate, we learn the cylindrical implicit interpolation weights (See Fig.~\ref{fig:CIIF}(b)):
\begin{equation}
\setlength{\abovedisplayskip}{8pt}
\setlength{\belowdisplayskip}{5pt}
w_{q,i} = g_{\theta}(z_i, (\alpha_i-\alpha_q, \beta_i - \beta_q)).
    \label{eq-CIIF}
\end{equation}

\begin{algorithm}[!t]
 \caption{The overall training process of our framework} 
 \label{alg} 
 \begin{algorithmic}[1]
     \STATE \textbf{Training data}: $I_{\mathrm{LR}},I_{\mathrm{HR}}^{\mathrm{GT}},D_{\mathrm{LR}}^{\mathrm{GT}}$
     \FOR{for i=1, $i\leq30$}
     \STATE \textbf{ODI SR task:}
     \STATE LR features $F_{\mathrm{LR}}^{\mathrm{RGB}}$ and $F_{\mathrm{LR}}^{\mathrm{U}}$ from $I_{\mathrm{LR}}$
     \STATE Up-sample $F_{\mathrm{LR}}^{\mathrm{RGB}}$ and $F_{\mathrm{LR}}^{\mathrm{U}}$ with CIIFs of parameters $g_\rho$ and $g_\theta$ to HR features $F_{\mathrm{HR}}^{\mathrm{RGB}}$ and $F_{\mathrm{HR}}^{\mathrm{U}}$, respectively
    \STATE SR RGB $\mu_{\mathrm{HR}}$ and SR uncertainty $\sigma_{\mathrm{HR}}$ from HR features, calculate Eq.~\eqref{eq2} \\
    \STATE \textbf{HR depth estimation task:}
      \STATE LR feature $F_{\mathrm{LR}}^{\mathrm{DE}}$ from $I_{\mathrm{LR}}$
      \STATE LR depth prediction $D_{\mathrm{LR}}$ from $F_{\mathrm{LR}}^{\mathrm{DE}}$, calculate Eq.~\eqref{eq-loss_lrde}
      \STATE \textbf{Share CIIF parameters $g_{\theta}$}
      \STATE Up-sample $F_{\mathrm{LR}}^{\mathrm{DE}}$ with CIIF of parameters $g_\theta$ to HR feature $F_{\mathrm{HR}}^{\mathrm{DE}}$
      \STATE HR depth prediction $D_{\mathrm{HR}}$ from $F_{\mathrm{HR}}^{\mathrm{DE}}$, calculate Eq.~\eqref{eq-loss_srde}
      \STATE \textbf{FD loss:}
      \STATE Transform $F_{\mathrm{HR}}^{\mathrm{RGB}}$ into $F_{\mathrm{HR}}^{\mathrm{T}}$, calculate Eq.~\eqref{eq14}
      \STATE HR Transformed depth $D_{\mathrm{HR}}^{\mathrm{T}}$ from $F_{\mathrm{HR}}^{\mathrm{T}}$, calculate Eq.~\eqref{eq-loss_trde}
      \STATE \textbf{Update framework} with Eq.~\eqref{eq-losstotal}
      
      \ENDFOR
 \end{algorithmic} 
\end{algorithm}

 After obtaining the interpolation weights $w_{q,i}$ in Eq.~\eqref{eq-CIIF}, we combine the weights $w_{q,i}$ with latent code $z_i$ from the LR feature $F_{\mathrm{LR}}$ to obtain an HR feature $F_{\mathrm{HR}}$. Finally, the HR features are processed with two different convolutional layers to output the super-resolved RGB image (\ie, mean) $\mu_{\mathrm{HR}}$ and super-resolved uncertainty map (\ie, variance) $\sigma_{\mathrm{HR}}$, respectively. 
% and HR uncertainty feature map, which are interpolated by $g_{\beta}$ and $g_{\theta}$, are decoded with two decoders to predict the high-resolution image $\mu_{\text{HR}}$ and high-resolution uncertainty $\sigma_{\text{HR}}$, respectively (See Fig.\ref{fig:framework}).
% Note that CIIF is not shared between mean features and variance features to encourage variance features to focus on structure information. 
Formally, by assuming that uncertainty meets the Laplace distribution, an HR ODI is described as:
\setlength{\abovedisplayskip}{5pt}
\setlength{\belowdisplayskip}{3pt}
\begin{equation}
    I_{\mathrm{HR}}=\mu_{\mathrm{HR}}+\epsilon \times \sigma_{\mathrm{HR}}, 
    \label{eq1}
\end{equation}
where $\epsilon$ represents the Laplace distribution with the zero mean and unit variance. Specifically, we employ the first stage of \cite{ning2021uncertainty} with the maximum likelihood estimation (MLE). For stable training, we utilize the logarithmic form of uncertainty $s=\textrm{log}\sigma_{\mathrm{HR}}$. The uncertainty-driven ODI SR loss function $\mathcal{L}_{\mathrm{UISR}}$ can be formulated as:

\setlength{\abovedisplayskip}{8pt}
\setlength{\belowdisplayskip}{3pt}
\begin{equation}
\mathcal{L}_{\mathrm{UISR}}=\frac{1}{N}\sum_{i=1}^{N}{e^{-s}}\left \| I^{\mathrm{GT},i}_{\mathrm{HR}}-\mu_{\mathrm{HR}}^{i}\right \|_1+2s,
    \label{eq2}
\end{equation}
\noindent where $N$ is the number of pixels, and $I^{\mathrm{GT}}_{\mathrm{HR}}$ is HR GT image. Except for constraining that the reconstructed high-resolution ODI should be similar to ground truth, $\mathcal{L}_{\mathrm{UISR}}$ additionally emphasizes high-frequency regions that exhibit high uncertainty. 

\subsection{Scene Structural Knowledge Transfer (SSKT)}
\label{CMKT}
\begin{figure}[t!]
    \centering
    \includegraphics[width=0.85\linewidth]{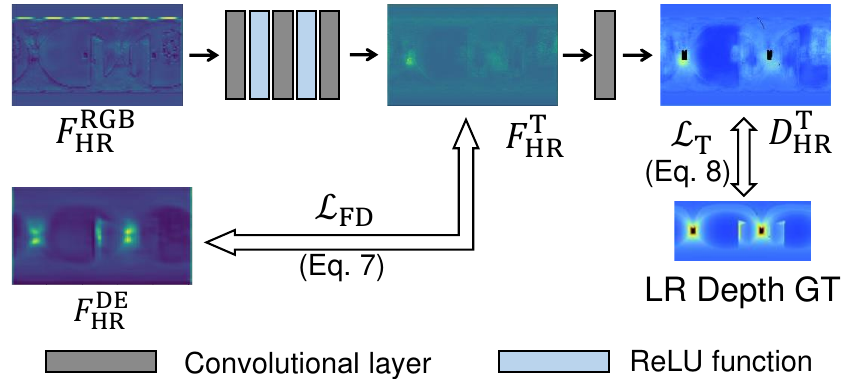}
    \centering
    \caption{An illustration of feature distillation (FD) loss.}
    \label{fig:fdloss}
    \vspace{-10pt}
\end{figure}

After extracting structural knowledge from the ODI SR task, we propose the SSKT module with two components to transfer the extracted knowledge to the HR depth estimation task, as shown in Fig.~\ref{fig:framework}. Buttressed by the learned interpolation weights of CIIF, we share the parameters $g_{\theta}$ of CIIF between the two tasks. This is because we think the learnable interpolation weights capture the relationships between pixels and thus represent the structural knowledge in both images and depth maps. Besides, to supplement high-resolution detailed features to the HR depth estimation task, we propose the feature distillation (FD) loss to transfer the scene structural knowledge from the ODI SR task, as shown in Fig.~\ref{fig:fdloss}. The detail of the CIIF sharing strategy and FD loss is introduced in the following part.

\noindent \textbf{CIIF sharing strategy:} 
The CIIF consists of the MLPs and is designed to predict the neural interpolation weights. As the interpolation weights for feature up-sampling can capture the correlations among neighboring pixel features, which can represent the scene structural knowledge.
Therefore, it is reasonable to share the parameters $g_{\theta}$ of CIIF between the two tasks for structural knowledge transfer. We do not consider sharing the parameters $g_{\rho}$, mainly because it additionally learns textural knowledge that will influence the performance of the HR depth estimation task (See Tab.~\ref{table:ablation-share}). Also, we design to directly share the parameters, instead of asynchronous methods~\cite{he2020momentum} to approximate the parameters of CIIFs. The reason is that the HR depth estimation task is in a weakly-supervised manner, where the structural details are scarcely possible to be recovered, while the ODI SR task can be fully-supervised to provide rich and fine scene structural details (See Tab.~\ref{table:ablation-interpolation}). Therefore, directly sharing parameters of CIIF between the two tasks is more effective for structural knowledge transfer.

\noindent \textbf{FD Loss:}
% We aim to help HR depth feature maps contain more structural details.
% It measures the correlation between 360◦ image modality and depth modality. 
The CIIF sharing strategy can benefit the feature up-sampling process in the HR depth estimation task. To further enhance the structural details in the HR depth prediction, we propose the FD loss as a structural regularization to help two tasks learn common scene structural knowledge. Although ODIs and their corresponding depth maps are two different representations of the same scene with high geometric similarity, simply applying feature similarity loss, \eg, cosine similarity loss, introduces unnecessary textural knowledge and thus destroys the smoothness of the depth map. As shown in Fig.~\ref{fig:fdloss}, we first transform the HR feature in the RGB part of the ODI SR network to simulate the HR feature in the HR depth estimation network. Given an HR feature $F^{\mathrm{RGB}}_{\mathrm{HR}}$ as input, an encoder consisting of three convolutional layers and ReLU activation functions is applied to obtain a transformed feature $F^{\mathrm{T}}_{\mathrm{HR}}$. We then aim to maximize the cosine similarity between $F^{\mathrm{T}}_{\mathrm{HR}}$ and HR feature $F^{\mathrm{DE}}_{\mathrm{HR}}$ from the HR depth estimation network. FD loss can be formulated as follows:
% As a result, $E^{tr}$ outputs $f^{tr}_{hr}$ as the depth feature which is compared with 
\setlength{\abovedisplayskip}{8pt}
\setlength{\belowdisplayskip}{5pt}
\begin{equation}
 \mathcal{L}_{\mathrm{FD}} = 1-\mathrm{cos}(F^{\mathrm{T}}_{\mathrm{HR}}, F^{\mathrm{DE}}_{\mathrm{HR}}).
\label{eq14}
\end{equation}

\noindent To supervise the transformed feature $F^{\mathrm{T}}_{\mathrm{HR}}$, we cascade a convolutional layer to process $F^{\mathrm{T}}_{\mathrm{HR}}$ and output HR transformed depth map $D^{\mathrm{T}}_{\mathrm{HR}}$. Then we down-sample $D^{\mathrm{T}}_{\mathrm{HR}}$ with the max-pooling operation $\mathrm{M}$ and supervise it using the LR depth GT map $D^{\mathrm{GT}}_{\mathrm{LR}}$ with Berhu loss $\mathcal{B}$~\cite{laina2016deeper}, formulated as follows:
\begin{equation}
    \begin{split}
        \mathcal{L}_{\mathrm{T}} &=\mathcal{B}(\mathrm{M}(D^{\mathrm{T}}_{\mathrm{HR}}) - D^{\mathrm{GT}}_{\mathrm{LR}}),\\
    \mathcal{B}(x)&=\left\{\begin{aligned}	
		&\left | x\right |, \left | x\right |\leq c\\
		&\frac{x^2+c^2}{2c}, \left | x\right | > c,
	\end{aligned}\right.
    \end{split}
    \label{eq-loss_trde}
\end{equation}
% \begin{align}
% \small
% &\mathcal{L}_{\mathrm{T}} =\mathcal{B}(\mathrm{M}(D^{\mathrm{T}}_{\mathrm{HR}}), D^{\mathrm{GT}}_{\mathrm{LR}}),\\
%     &\mathcal{B}(x)=\left\{\begin{aligned}	
% 		&\left | x\right |, \left | x\right |\leq c\\
% 		&\frac{x^2+c^2}{2c}, \left | x\right | > c.
% 	\end{aligned}\right.
%  \label{eq-loss_trde}
% \end{align}
where c is a threshold and set to 0.2 empirically~\cite{Jiang2021UniFuseUF}.
\subsection{HR Depth Estimation}
\label{DESR}
Lastly, the HR depth estimation network leverages the transferred scene structural knowledge to estimate the HR depth map. As shown in Fig.~\ref{fig:framework}, 
taking an LR ODI $I_{\mathrm{LR}}$ as input, the feature extractor outputs an LR feature $F^{\mathrm{DE}}_{\mathrm{LR}}$ with the same resolution as $I_{\mathrm{LR}}$. Then, the LR feature $I_{\mathrm{LR}}$ is processed with two paths. First, $I_{\mathrm{LR}}$ is processed with a convolutional layer to output an LR depth prediction $D_{\mathrm{LR}}$, which is directly supervised by $D_{\mathrm{LR}}^{\mathrm{GT}}$ through Berhu loss $\mathcal{B}$ as follows:
\setlength{\abovedisplayskip}{8pt}
\setlength{\belowdisplayskip}{5pt}
\begin{equation}
 \mathcal{L}_{\mathrm{LD}} = \mathcal{B}(D_{\mathrm{LR}}, D^{\mathrm{GT}}_{\mathrm{LR}}).
\label{eq-loss_lrde}
\end{equation}

Meanwhile, $F^{\mathrm{DE}}_{\mathrm{LR}}$ is up-sampled with CIIF of shared parameters $g_\theta$, to generate an HR feature $F^{\mathrm{DE}}_{\mathrm{HR}}$. $F^{\mathrm{DE}}_{\mathrm{HR}}$ is then fed into another convolutional layer to predict an HR depth map $D_{\mathrm{HR}}$. We first down-sample $D_{\mathrm{HR}}$ with the max-pooling operation $\mathrm{M}$ and then apply Berhu loss to supervise it as follows:
\setlength{\abovedisplayskip}{8pt}
\setlength{\belowdisplayskip}{5pt}
\begin{equation}
 \mathcal{L}_{\mathrm{HD}} = \mathcal{B}(\mathrm{M}(D_{\mathrm{HR}}), D^{\mathrm{GT}}_{\mathrm{LR}}).
\label{eq-loss_srde}
\end{equation}

\subsection{Objective Function}

The total loss $\mathcal{L}_{\mathrm{TOTAL}}$ includes five parts: uncertainty-driven ODI SR loss $\mathcal{L}_{\mathrm{UISR}}$, FD loss $\mathcal{L}_{\mathrm{FD}}$, and three weakly-supervised depth estimation losses: LR depth estimation loss $\mathcal{L}_{\mathrm{LD}}$, HR depth estimation loss $\mathcal{L}_{\mathrm{HD}}$, and transformed depth estimation loss $\mathcal{L}_{\mathrm{T}}$. $\mathcal{L}_{\mathrm{TOTAL}}$ can be formulated as follows:

\begin{equation}
\begin{split}
\mathcal{L}_{\mathrm{TOTAL}} =  \lambda_1 \mathcal{L}_{\mathrm{UISR}} +  \lambda_2 \mathcal{L}_{\mathrm{FD}} &+ \lambda_3 \mathcal{L}_{\mathrm{LD}} \\
+ \lambda_4 \mathcal{L}_{\mathrm{HD}} + \lambda_5 \mathcal{L}_{\mathrm{T}},
% + \lambda_4  \mathcal{L}_{ws} + \lambda_5  \mathcal{L}_{edge}.
\label{eq-losstotal}
\end{split}
\end{equation}
where $\lambda_i, i\in \{1,2,3,4,5\}$ is a hyper-parameter for $i$-th loss term in Eq.~\eqref{eq-losstotal}.

% \begin{table}[t!]
% \renewcommand{\arraystretch}{1.5}
% \setlength{\tabcolsep}{1.2mm}
% \caption{The number of samples in training set, validation set, and testing set in three datasets.}
%   \centering
%   {
%   \begin{tabular}{c | c | c | c}
%   \hline
%  ~ & \multicolumn{3}{c}{Dataset} \\
%   \cline{2-4}
%   ~ & Stanford2D3D~\cite{Armeni2017Joint2D} & Matterport3D~\cite{chang2017matterport3d}& 3D60~\cite{Zioulis2018OmniDepthDD} \\
%   \hline
%   Training & 1000 & 7829 & -- \\
%   \hline
%   Validation & 40 & 947 & -- \\
%   \hline
%   Testing & 373 & 2014 & 1298 \\
%   \hline
%   \end{tabular}}
% \label{table:datasets}
% \end{table}

\begin{table*}[t]
\caption{Quantitative comparison with fully-supervised methods (with HR depth GT) on Stanford2D3D and Matterport3D dataset.}
\centering
\footnotesize
\renewcommand{\arraystretch}{1.2}
\makebox[\textwidth]{
\begin{tabular}{c | c | c | c | c | c | c | c | c | c}
\hline
Dataset & Backbone & Inp. size & Supervision & MAE $\downarrow$ & Abs Rel $\downarrow$ & RMSE log $\downarrow$ & $\delta_1 \uparrow$ & $\delta_2 \uparrow$ & $\delta_3 \uparrow$ \\
\hline
\hline
\multirow{12}{*}{Stanford2D3D~\cite{Armeni2017Joint2D}} & \multirow{4}{*}{OmniDepth~\cite{Zioulis2018OmniDepthDD}} & \multirow{2}{*}{128$\times$256 ($\times4$)} & Fully & 0.3248 & 0.1653 & 0.1065 & 77.68 & 91.71 & 96.54 \\
~ & ~  & ~ & Weakly (Ours) & \textbf{0.3126} & \textbf{0.1566} & \textbf{0.1043} & \textbf{78.87} & \textbf{92.27} & \textbf{97.03}  \\
\cline{3-10}
~ & ~  & \multirow{2}{*}{64$\times$128 ($\times8$)} & Fully & 0.3484 & 0.1830 & \textbf{0.1117} & 75.71 & \textbf{90.90} & 96.13  \\
~ & ~ & ~ & Weakly (Ours) & \textbf{0.3480} & \textbf{0.1772} & 0.1154 & \textbf{75.85} & 90.59 & \textbf{96.19}  \\
\cline{2-10}
~ & \multirow{4}{*}{UniFuse-Fusion~\cite{Jiang2021UniFuseUF}} & \multirow{2}{*}{128$\times$256 ($\times4$)} & Fully & 0.2785 & 0.1409 & \textbf{0.0868} & \textbf{84.01} & \textbf{95.31} & \textbf{98.45} \\
~ & ~ & ~ & Weakly (Ours) & \textbf{0.2784} & \textbf{0.1398} & 0.0929 & 82.23 & 94.20 & 97.83 \\
\cline{3-10}
~ & ~  & \multirow{2}{*}{64$\times$128 ($\times8$)} & Fully & \textbf{0.3124} & \textbf{0.1582} & \textbf{0.0975} & \textbf{80.16} & \textbf{93.03} & \textbf{97.46} \\
~ & ~ & ~ & Weakly (Ours) & 0.3135 & 0.1585 & 0.1037 & 78.06 & 92.71 & 97.37 \\
\cline{2-10}
~ & \multirow{4}{*}{UniFuse-Equi~\cite{Jiang2021UniFuseUF}}  & \multirow{2}{*}{128$\times$256 ($\times4$)} & Fully & \textbf{0.2732} & 0.1403 & \textbf{0.0870} & \textbf{83.70} & \textbf{94.90} & 98.24 \\
~ & ~ & ~ & Weakly (Ours) & 0.2747 & \textbf{0.1403} & 0.0908 & 83.13 & 94.83 & \textbf{98.25} \\
\cline{3-10}
~ & ~  & \multirow{2}{*}{64$\times$128 ($\times8$)} & Fully & 0.3263 & \textbf{0.1619} & \textbf{0.1011} & 77.37 & \textbf{92.94} & \textbf{97.33} \\
~ & ~ & ~ & Weakly (Ours) & \textbf{0.3241} & 0.1630 & 0.1040 & \textbf{78.69} & 92.62 & 97.21 \\
\hline
\multirow{10}{*}{Matterport3D~\cite{chang2017matterport3d}} & \multirow{6}{*}{UniFuse-Fusion~\cite{Jiang2021UniFuseUF}} & \multirow{2}{*}{128$\times$256 ($\times4$)} & Fully & 0.3890 & \textbf{0.1547}  & \textbf{0.0966} & \textbf{79.11} & 92.99 & 96.97 \\
~ & ~  & ~ & Weakly (Ours) & \textbf{0.3827} & 0.1569 & 0.0986 & 78.75 & \textbf{93.19} & \textbf{97.27}  \\
\cline{3-10}
~ & ~ & \multirow{2}{*}{64$\times$128 ($\times8$)} & Fully & 0.4779  & \textbf{0.1973} & \textbf{0.1179} & \textbf{70.90} & 88.98 & 95.29 \\
~ & ~ & ~ & Weakly (Ours) & \textbf{0.4767} & 0.2025 & 0.1220 & 69.80 & \textbf{89.21} & \textbf{95.57} \\
\cline{3-10}
~ & ~ & \multirow{2}{*}{256$\times$512 ($\times4$)} & Fully &  0.3233 & 0.1314  & \textbf{0.0811} & 84.39 & 95.15 & \textbf{98.01} \\
~ & ~ & ~ & Weakly (Ours) & \textbf{0.3152} & \textbf{0.1264}  & 0.0837 & \textbf{85.04} & \textbf{95.30} & 97.99 \\
\cline{2-10}
~ & \multirow{4}{*}{OmniDepth~\cite{Zioulis2018OmniDepthDD}} & \multirow{2}{*}{128$\times$256 ($\times4$)} & Fully & 0.4483 & 0.1914 & 0.1127 & 72.95 & 90.21 & 95.75 \\
~ & ~  & ~ & Weakly (Ours) & \textbf{0.4237}
& \textbf{0.1828} & \textbf{0.1082} & \textbf{74.44} & \textbf{91.34} & \textbf{96.51} \\
\cline{3-10}
~ & ~ & \multirow{2}{*}{64$\times$128 ($\times8$)} & Fully & \textbf{0.4823}  & 0.2127 & \textbf{0.1209} & \textbf{69.80} & 88.68 & \textbf{95.38} \\
~ & ~ & ~ & Weakly (Ours) & 0.4843 & \textbf{0.2076} & 0.1219 & 69.76 & \textbf{88.70} & 95.20 \\
% \cline{2-10}
% ~ & \multirow{2}{*}{360MonoDepth~\cite{ReyArea2021360MonoDepthH3}} & 128$\times$256 ($\times4$) & - &  &  &  &  &  & \\
% ~ & ~ & 256$\times$512 ($\times4$) & - &  &  &  &  &  & \\
\hline
\end{tabular}}
\label{table:comparison}
\end{table*}

\section{Experiments}

\begin{figure*}[t!]
    \centering
    \includegraphics[width=0.75\textwidth]{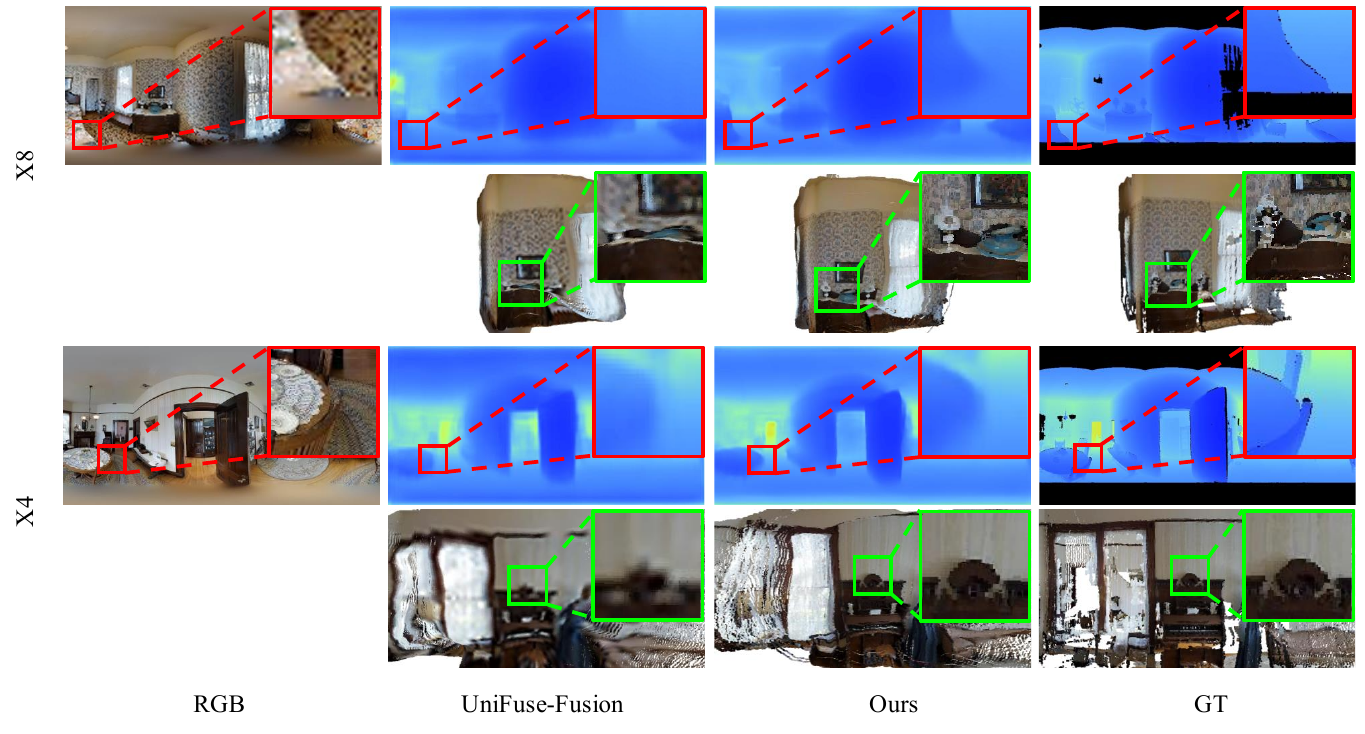}
    \caption{Visual comparison of fully-supervised method (\ie, UniFuse-Fusion) and ours on Matterport3D dataset. Best viewed in color.}
    \label{fig:compare_matterport}
    \vspace{-15pt}
\end{figure*}

% \begin{figure*}[t!]
%     \centering
%     \includegraphics[width=0.82\textwidth]{Figure/com_matter.pdf}
%     \caption{\textbf{Visual comparison of fully-supervised methods and ours} on Matterport3D dataset. Best viewed in color.}
%     \label{fig:compare_matterport}
% \end{figure*}

\subsection{Implementation Details}

\noindent\textbf{Dataset.} We conduct on three datasets: Stanford2D3D~\cite{Armeni2017Joint2D}, Matterport3D~\cite{chang2017matterport3d} and 3D60 \cite{Zioulis2018OmniDepthDD}. Stanford2D3D and Matterport3D are real-world datasets, and 3D60 is a synthetic dataset, provided by OmniDepth~\cite{Zioulis2018OmniDepthDD}. 
% rendered using two realistic datasets and two synthetic datasets.
We follow UniFuse to split the three datasets respectively. The sizes of ODIs and paired depth maps are 512$\times$1024 in Stanford2D3D and Matterport3D and 256$\times$512 in 3D60.
On real-world datasets, the up-sampling factors are set to $\times4$ and $\times8$. We train, validate, and test on the two real-world datasets, which are more challenging and have higher resolution than the synthetic dataset. In addition, we test on the synthetic dataset with $\times8$ up-sampling factor to verify the generalization ability.

\noindent\textbf{Implementation.} We use three omnidirectional depth estimation methods as backbones to extract features from the input ODIs: OmniDepth~\cite{Zioulis2018OmniDepthDD} with spherical convolutions, UniFuse-Fusion~\cite{Jiang2021UniFuseUF} with ERP type images and cube maps, and light-weight UniFuse-Equi~\cite{Jiang2021UniFuseUF} with only ERP type images. As for the feature extractor of the ODI SR task, we directly employ the commonly-used EDSR-baseline~\cite{lim2017enhanced} to extract features from the input LR ODIs. We down-sample ODIs with Bicubic interpolation, while down-sample omnidirectional depth maps with Nearest interpolation. During training, we augment data with horizontal flipping and rolling. We collaboratively train the two tasks for 30 epochs using the Adam optimizer~\cite{kingma2014adam} with batch size 2. The initial learning rate is set to $1e-4$, and reduced by 2 times per 5 epochs. For hyper-parameters in $\mathcal{L}_{\mathrm{TOTAL}}$, we empirically set $\lambda_1=\lambda_2=10 $, and $\lambda_3=\lambda_4=\lambda_5=1$. For a fair comparison, we re-train the compared methods based on their provided settings. 

\subsection{Quantitative and Qualitative Evaluation}
We evaluate our method with standard metrics, including mean absolute error (MAE), absolute relative error (Abs Rel), and root mean square error in log space (RMSE log). We also utilize three accuracy measures ($\delta_i$), which are defined as the fraction of pixels where the relative error between the estimated depth and ground-truth depth is less than the threshold $i$, $i \in \{1.25, 1.25^2, 1.25^3\}$. We only measure on valid pixels in the ground-truth depth map. 

\begin{figure*}[!t]
    \centering
    \includegraphics[width=0.85\linewidth]{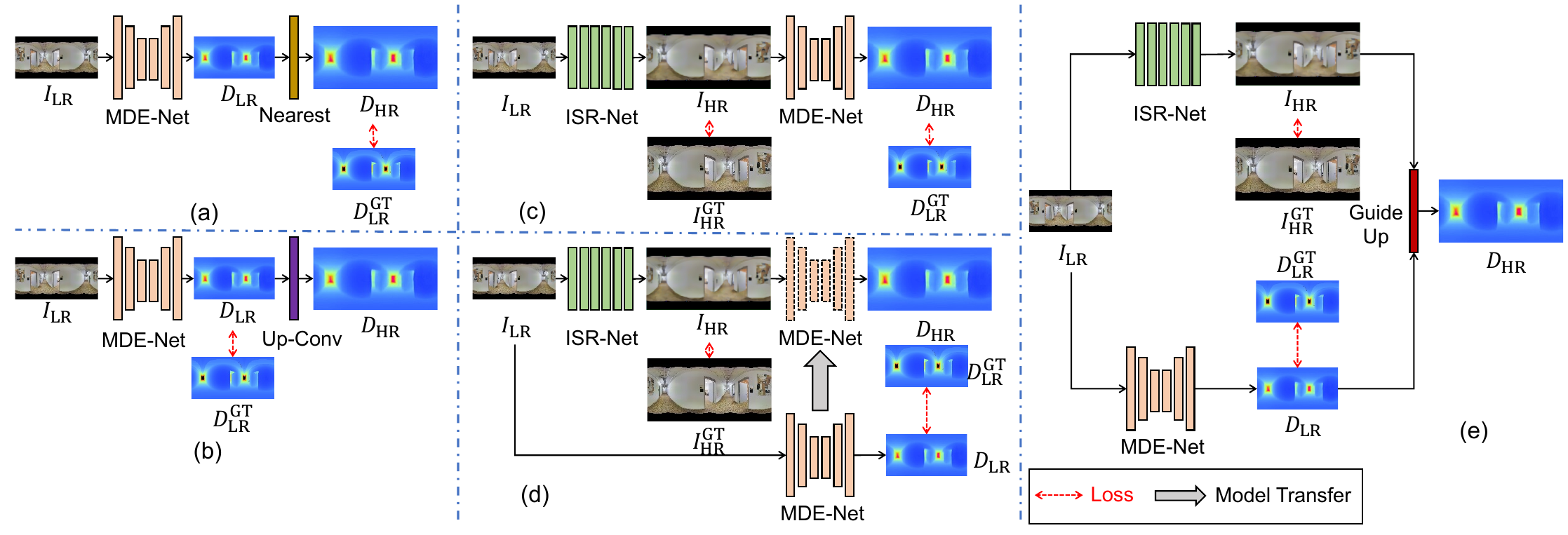}
    \caption{The pipelines of baseline methods with five different settings. To be consistent with our weakly-supervised setting, these baseline methods only utilize LR ODIs as inputs and LR depth GT maps for supervision. MDE-Net means the monocular depth estimation network, ISR-Net means the ODI super-resolution network, Up-Conv represents the learnable up-sampling convolutional layer, and Guide-Up represents guided image filter~\cite{he2012guided}.}
    \label{fig:setting}
    \vspace{-10pt}
\end{figure*}

% Our method restricts the input resolution, and only utilizes LR depth ground truth for supervision. 

\noindent \textbf{Fully-supervised methods.} We compare with fully-supervised methods that utilize HR depth GT maps. To the best of our knowledge, there are no existing methods designed to generate HR depth prediction from an LR ODI. To this end, we add learnable up-sampling convolutional layers to OmniDepth~\cite{Zioulis2018OmniDepthDD}, UniFuse-Fusion~\cite{Jiang2021UniFuseUF}, and UniFuse-Equi~\cite{Jiang2021UniFuseUF} and re-train them with HR depth GT maps. As shown in Tab.~\ref{table:comparison}, our weakly-supervised method not only achieves comparable results with corresponding fully-supervised methods (\eg, 3rd-4th rows) but also outperforms corresponding supervised methods in some cases (\eg, 1st-2nd rows). For example, in the 4th row of Fig.~\ref{fig:compare_matterport}, our method predicts the furniture with regular shapes, while the other corresponding fully-supervised methods predict the shape with more or less warp.

The adaptability of our method in different input sizes is also verified, \eg, $128\times256$ and $64\times128$. Our weakly-supervised method outperforms the fully-supervised methods in several metrics, \eg, MAE, Abs Rel, and $\delta_3$. For example, with OmniDepth as the backbone with $128\times256$ input size, our method obtains $3.8\%$ gain in MAE metric and $5.3\%$ gain in Abs Rel metric. When choosing a smaller input size $64\times128$, our method has inferior or comparable results compared with the fully-supervised methods. By taking $\times 8$ up-sampling factor and UniFuse-Fusion as the backbone, our method has no metric outperforming the fully supervised method on the Stanford2D3D dataset. We think it is because the LR depth GT map is too sparse, which only contains 1/64 supervision signals of an HR depth GT map. 

% can achieve similar results compared to the fully-supervised method, even though the LR depth GT map only contains 1/64 supervision signals of an HR depth GT map. 

In addition, we train on the Matterport3D dataset with UniFuse-Fusion and OmniDepth as backbone and achieve comparable results with the corresponding fully-supervised methods. For example, with UniFuse-Fusion as the backbone and $\times8$ up-sampling factor, our method improves $1.6\%$ in the MAE metric and $0.3\%$ in $\delta_3$ metric. Finally, in Tab.~\ref{table:comparison}, we verify the effectiveness of our method with higher resolution, \ie, $1024\times2048$. Given the input resolution is $256\times512$ and the up-sampling factor is 4, the results show that our weakly-supervised method performs well. For example, compared with the fully-supervised method, ours can improve 0.0050 in Abs Rel.

% In total, our weakly-supervised can adapt to various backbones, input sizes, and datasets, which demonstrates its strong robustness under different occasions. Our method can also easily be applied to higher resolution, such as $1024\times2048$. We contribute the effectiveness to the proposed SSKT module, which successfully transfers essential scene structural knowledge from the ODI SR task to HR depth estimation task. Importantly, we prove that the way of transferring the scene structural knowledge from the HR image modality and corresponding LR depth map is feasible to achieve the goal of HR depth estimation without any extra inference cost.

\begin{figure}[t!]
    \centering
    \includegraphics[width=0.85\linewidth]{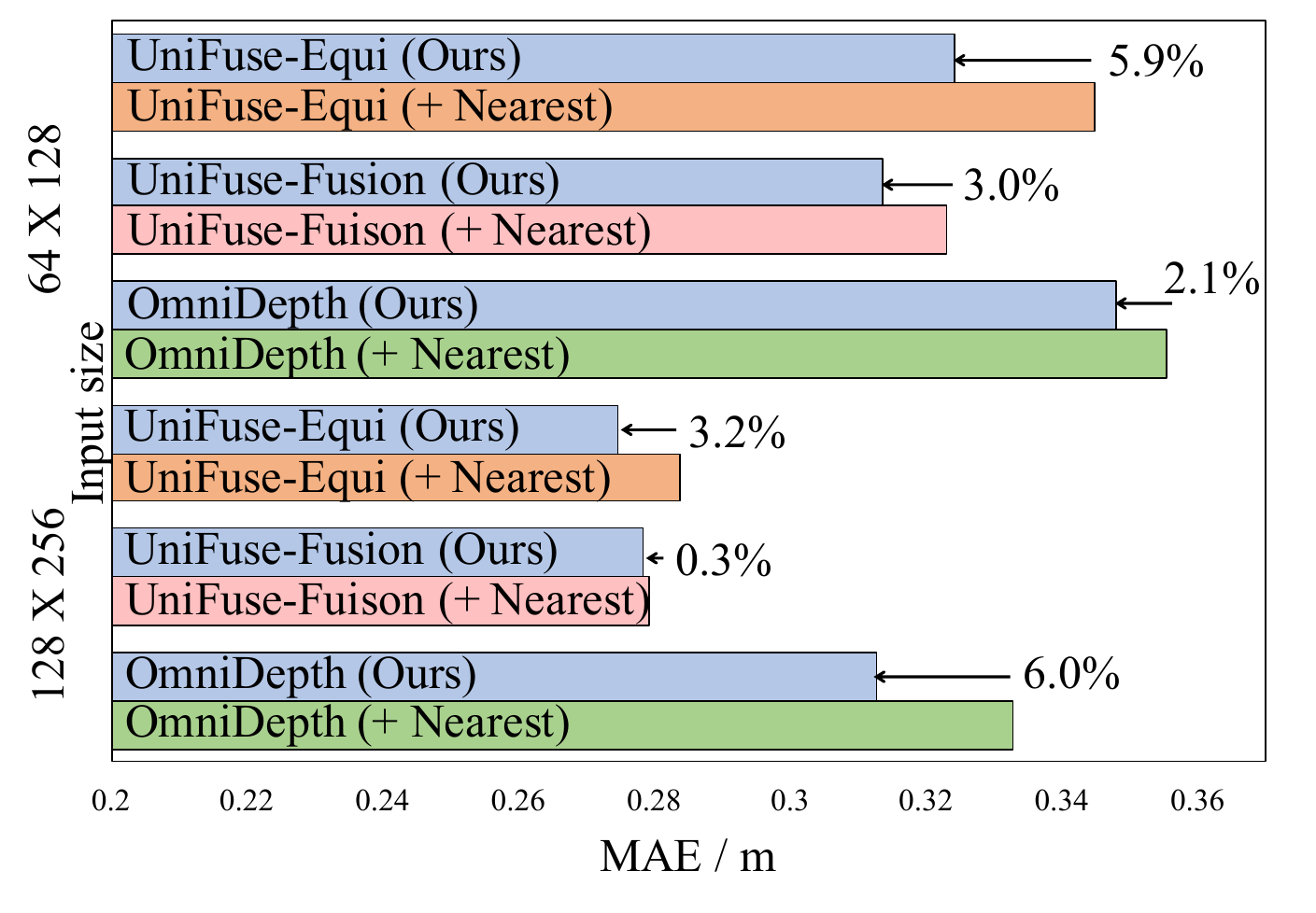}
    \vspace{-15pt}
    \caption{Quantitative comparison with the baseline methods in the \textbf{ Setting 1} with different up-sampling factors and backbones on Stanford2D3D dataset.}
    \label{fig:baseline}
\end{figure}

\begin{table}[t]
\caption{Comparison with the baseline methods that use LR ODIs as inputs, and LR depth ground truth maps for supervision.}
\renewcommand{\arraystretch}{1.2}
\setlength{\tabcolsep}{4.2mm}
  \centering
  {
  \begin{tabular}{c | c c c}
  \hline
  & Abs Rel $\downarrow$ & RMSE log $\downarrow$ & $\delta_3 \uparrow$ \\
  \hline
   Setting 2 & 0.1830 & 0.1180 & 95.84 \\
   Setting 3 & 0.3848 & 0.2847 & 76.94 \\
   Setting 4 & 0.4523 & 0.3849 & 58.17 \\
    Setting 5 & 0.4422 & 0.2617 & 78.50 \\
  Ours & \textbf{0.1585} & \textbf{0.1037} & \textbf{97.37} \\
  \hline
  \end{tabular}}
\label{table:ablation-SRDE}
\vspace{-10pt}
\end{table}

\noindent \textbf{Baseline methods.} There are various ways to achieve HR omnidirectional depth estimation using an LR ODI as the input and an LR depth map for supervision. To fully verify the effectiveness of our framework, we compare our method with all possible baseline methods.

\noindent \textbf{1) Setting 1:} As shown in Fig.~\ref{fig:setting}(a), it is to estimate LR depth map from LR input ODI, and directly up-sample the LR depth prediction with Nearest interpolation. As shown in Fig.~\ref{fig:baseline}, our method outperforms baseline methods with Nearest interpolation in different input sizes and backbones. For example, with OmniDepth as the backbone, our method leads to a 6.0\% improvement in the MAE metric, which proves the effects of structural knowledge transfer from the ODI SR task. 

\noindent \textbf{2) Setting 2:} In Fig.~\ref{fig:setting}(b), it is similar to Setting 1, except for replacing the Nearest interpolation with learnable up-sampling convolutional layers. The result of Setting 2 is 0.1830 Abs Rel metric, which is lower than our method (0.1585 Abs Rel). We analyze the reason that the additional up-sampling convolutional layer can not be fully supervised by HR depth GT maps, causing its performance to drop greatly.

\noindent \textbf{3) Setting 3:} In Fig.~\ref{fig:setting}(c), it is to first up-sample LR ODIs with an ODI SR network, \ie, EDSR-baseline~\cite{lim2017enhanced}, which is fully-supervised with HR ground truth ODIs. Then, the super-resolved ODI is fed into a depth estimation network to generate an HR depth prediction. The HR depth prediction is down-sampled with the Nearest interpolation to be compared with the LR depth GT map. In Tab.~\ref{table:ablation-SRDE}, setting 3 performs badly, due to the super-resolved ODI containing much noise, which disturbs the following depth estimation network.

\noindent \textbf{4) Setting 4:} In Fig.~\ref{fig:setting}(d), it is similar to Setting 3, except that the depth estimation network is trained with LR ODI as input. We then directly transfer the pre-trained depth estimation network into the HR domain with the super-resolved ODI as input. However, it only obtains 0.4523 Abs Rel, which is worse than Setting 3 and our method. Its main drawback is that during training, the depth estimation network has no access to HR ODIs or HR depth maps. 

\begin{table}[t]
\caption{Ablation studies on different 
\textbf{sharing strategies}.}
\setlength{\tabcolsep}{1mm}
\renewcommand{\arraystretch}{1.2}
 \centering
  {
  \begin{tabular}{c c c | c c}
  \hline
  RGB (Mean) & Uncertainty (Variance) & Depth & MAE $\downarrow$ & Abs Rel $\downarrow$\\
  \hline
%   & & & 0.3323 & 0.1653 \\
 \checkmark & --- & \checkmark & 0.3225 & 0.1708 \\
  \checkmark & & \checkmark & 0.3179 & 0.1588 \\
  & \checkmark & \checkmark & \textbf{0.3135} & \textbf{0.1585} \\
  \checkmark & \checkmark & \checkmark & 0.3234 & 0.1686 \\
  \hline
  \end{tabular}}
\label{table:ablation-share}
\vspace{-5pt}
\end{table}

\noindent \textbf{5) Setting 5:} In Fig.~\ref{fig:setting}(e), it first predicts a super-resolved image with an ODI SR network, \ie, EDSR-baseline~\cite{lim2017enhanced}, and an LR depth prediction with a depth estimation network. Then, it utilizes the super-resolved ODI to guide the interpolation process of the LR depth prediction with guided image filter~\cite{he2012guided}. However, the result is similar to that of Setting 4, due to the noise contained in the super-resolved image.

\subsection{Ablation Studies}
 We conduct on the Stanford2D3D dataset with $\times8$ factor.

% \begin{figure*}[!t]
%     \centering
%     \includegraphics[width=0.95\linewidth]{Figure/setting.pdf}
%     \caption{The pipelines of baseline methods with five different settings. To be consistent with our weakly-supervised setting, these baseline methods only utilize LR ODIs as inputs and LR depth GT maps for supervision. MDE Net means the monocular depth estimation network, ISR Net means the ODI super-resolution network, Up Conv represents the learnable up-sampling convolutional layer, and Guide Up represents guided image filter~\cite{he2012guided}.}
%     \label{fig:setting}
% \end{figure*}

% \begin{figure*}[!t]
%     \centering
%     \includegraphics[width=0.95\linewidth]{Figure/ab_stanford.pdf}
%     \caption{\textbf{Visual comparison of the impacts of uncertainty estimation in the ODI SR task and FD loss} on the overall framework.}
%     \label{fig:ablation}
% \end{figure*}

\begin{table}
\caption{Ablation studies on CIIF about whether learning interpolation weights, parameter update, and the type of coordinate system.}
\setlength{\tabcolsep}{1.2mm}
\renewcommand{\arraystretch}{1.2}
  \centering
  {
  \begin{tabular}{c|c|c|c c c}
  \hline
 ~ & Update & Coordinate & MAE $\downarrow$ & Abs Rel $\downarrow$ & $\delta_1 \uparrow$ \\
  \hline
  Bicubic & - & - & 0.3209 & 0.1626 & 77.92 \\
  LIIF & Synchronous & Spherical & 0.3302 & 0.1690 & 76.36 \\
  CIIF & Asynchronous & Cylindrical & 0.3329 & 0.1692 & 76.94 \\
  CIIF & Synchronous & Cylindrical & \textbf{0.3135} & \textbf{0.1585} & \textbf{78.06}\\
  \hline
  \end{tabular}}
\label{table:ablation-interpolation}
\vspace{-10pt}
\end{table}

% \begin{table}[t]
% \caption{Ablation studies on FD loss about different calculations of feature similarity and whether applying transform.}
% \setlength{\tabcolsep}{3mm}
% \renewcommand{\arraystretch}{1.2}
%  \centering
%   {
%   \begin{tabular}{c | c | c c}
%   \hline
%    & Transform & Abs Rel $\downarrow$ & $\delta_3 \uparrow$\\
%   \hline
%   \multirow{2}{*}{Feature affinity~\cite{Wang2020DualSL}} &  & 0.1645  & 96.93  \\ 
%   ~ & \checkmark & 0.1630 & \textbf{97.42} \\ \hline
%    \multirow{2}{*}{Cosine similarity} & & 0.1723 & 97.34 \\
%    ~ & \checkmark & \textbf{0.1585} & 97.37 \\
%   \hline
%   \end{tabular}}
% \label{table:ablation-fd}
% \end{table}

\noindent \textbf{Uncertainty estimation.}
% Tab.~\ref{table:ablation-uncertainty} shows the influence of uncertainty estimation.
In the first row of Tab.~\ref{table:ablation-share}, we first omit uncertainty estimation in the ODI SR task and share the parameters of CIIF between feature up-sampling processes in the ODI SR task and HR depth estimation task. Removing uncertainty estimation leads to 2.8\% MAE and 7.2\% Abs Rel reduction. Directly transferring knowledge between the ODI SR task and the HR depth estimation task obtains sub-optimal performance, because the redundant textural details in the ODI SR task might interfere with the HR depth estimation task.

\noindent \textbf{Sharing strategy of CIIF.} To verify the effectiveness of our proposed sharing strategy, we additionally experiment on other combinations in Tab.~\ref{table:ablation-share}. Specifically, in the second row, we choose to share the parameters of CIIF between feature up-sampling processes in the RGB part of the ODI SR task, and the HR depth estimation task. Moreover, we study the influence of sharing the parameters among the feature up-sampling processes in the RGB part of the ODI SR task, the uncertainty part of the ODI SR task, and the HR depth estimation task. However, we find that Abs Rel degrades 6.0\% compared with sharing between the uncertainty part and HR depth estimation task. It also proves that textural details of directly transferring knowledge from features in the RGB part of the ODI SR task can cause negative effects on the structural knowledge learning of the HR depth estimation task.

\noindent \textbf{Components of CIIF.} We replace CIIF with Bicubic interpolation followed by a convolution layer, which is commonly used in ~\cite{Jiang2021UniFuseUF, li2022omnifusion}. In Tab.~\ref{table:ablation-interpolation}, we find that CIIF gains 2.5\% in Abs Rel. We ascribe it as learning correlation among adjacent pixels is more effective for structural knowledge learning and transferring between different modals.

The parameters of CIIF are all contained in an MLP. We share the parameters of CIIF between the ODI SR task and the HR depth estimation task. To verify it, we conduct a momentum update~\cite{he2020momentum} for the HR depth estimation task, which is considered as an asynchronous update. From Tab.~\ref{table:ablation-interpolation}, asynchronous update reduces performance greatly. We think the reason is that scene structural knowledge is essential for the HR depth estimation task, and asynchronous update destroys the common structural representations between the two tasks.

In addition, we investigate the effectiveness of cylindrical representation. As shown in Tab.~\ref{table:ablation-interpolation}, exploiting cylindrical representation obtains a 5.1\% gain in MAE metric compared with utilizing spherical angle coordinates. Note that using spherical angle coordinates is equal to using planar pixel coordinates in LIIF~\cite{chen2021learning, ai2022deep}. The result demonstrates that the cylindrical coordinate system is more proper to describe the spatial correlation of neighboring pixels in ERP type ODIs.

\begin{table}
\caption{Ablation studies on loss functions.}
\setlength{\tabcolsep}{3.2mm}
  \centering
  \renewcommand{\arraystretch}{1.2}
  {
  \begin{tabular}{c c | c | c c}
  \hline
$\mathcal{L}_{\mathrm{LD}}$ & $\mathcal{L}_{\mathrm{FD}}$ & Transform & Abs Rel $\downarrow$ & $\delta_3 \uparrow$ \\
  \hline
 & \checkmark & \checkmark & 0.1642 & 97.30 \\
\checkmark & &  & 0.1655 & 97.10 \\
\checkmark & \checkmark  &  & 0.1723 & 97.34 \\
\checkmark & \checkmark & \checkmark & \textbf{0.1585} & \textbf{97.37} \\
  \hline
  \end{tabular}}
\label{table:loss}
\end{table}

\begin{table}[!t]
\caption{Comparison of estimating uncertainty (Ours) with estimating the gradient for the ODI SR task.}
\setlength{\tabcolsep}{3.0mm}
  \centering
  \renewcommand{\arraystretch}{1.2}
  {
  \begin{tabular}{c | c c c}
  \hline
   & MAE $\downarrow$ & Abs Rel $\downarrow$ & $\delta_1 \uparrow$ \\
  \hline
  Gradient & 0.3141 & 0.1635 & \textbf{78.30} \\
 Uncertainty (Ours) & \textbf{0.3135} & \textbf{0.1585} & 78.06 \\
  \hline
  \end{tabular}}
\label{table:ablation-uncertainty}
\vspace{-10pt}
\end{table}

\noindent \textbf{Loss function.} By default, $\mathcal{L}_{\mathrm{UISR}}$ and $\mathcal{L}_{\mathrm{HD}}$ are necessary for the training. Therefore, we investigate the effectiveness of $\mathcal{L}_{\mathrm{LD}}$ and $\mathcal{L}_{\mathrm{FD}}$. In Tab.~\ref{table:loss}, without $\mathcal{L}_{\mathrm{LD}}$, the performance degrades 3.5\% in Abs Rel metric. It demonstrates that $\mathcal{L}_{\mathrm{LD}}$ can utilize the LR depth GT maps to help obtain more accurate LR features for the HR depth estimation task. Without $\mathcal{L}_{FD}$, the performance also degrades 1.6\% in the RMSE log metric. In addition, we omit the transformation process and find that the Abs Rel metric degrades from 0.1585 to 0.1723.

% For FD loss, we further investigate the choice of different calculations of feature similarity in Tab.~\ref{table:ablation-fd}. Without transform, the feature affinity loss~\cite{Wang2020DualSL} even performs worse than cosine-similarity. After adding the transform process, the feature affinity loss improves in both metrics, such as 0.49\% in $\delta_3$ metric. The results show that the transform process is more important than the selection of the calculation of feature similarity.

% with FD loss, our weakly-supervised method can obtain 4.2\% Abs Rel and 1.6\% gains. It shows that transferring knowledge from HR mean features can supplement necessary structural information for depth branch.

\subsection{Discussion}

\noindent \textbf{Structural representation.} To investigate the form of structural representation, we replace uncertainty estimation with gradient estimation~\cite{ma2020structure}. In Tab.~\ref{table:ablation-uncertainty}, uncertainty estimation performs better than gradient estimation. 
% We think the reason is that structural information is consistent between image and depth modalities, and extracting structural information can effectively filter color and texture information from the mean branch. 
Compared with the gradient that only matches fixed edges of ODIs with the certain data distribution~\cite{ning2021uncertainty}, uncertainty estimation is from the data itself and is more flexible for learning structural knowledge.

\begin{table}[t]
\caption{Discussion about employing different backbones for the ODI SR task.}
\renewcommand{\arraystretch}{1.2}
\setlength{\tabcolsep}{6mm}
  \centering
  {
  \begin{tabular}{c | c c}
  \hline
  & Abs Rel $\downarrow$  & $\delta_3 \uparrow$ \\
  \hline
   SwinIR~\cite{liang2021swinir}  & 0.1673  &  96.99 \\
   Omni-SR~\cite{wang2023omni} & 0.1733 & 96.86 \\
   EDSR-baseline~\cite{lim2017enhanced} & \textbf{0.1585} & \textbf{97.37} \\
  \hline
  \end{tabular}}
\label{table:srbackbone}
\end{table}

\begin{table}[!t]
\caption{Comparison of generalization capability.}
\setlength{\tabcolsep}{3mm}
  \centering
  \renewcommand{\arraystretch}{1.2}
  {
  \begin{tabular}{c|c | c c c}
  \hline
   & Method & MAE $\downarrow$ & $\delta_3 \uparrow$ \\
  \hline
  Stanford2D3D & UniFuse-Fusion~\cite{Jiang2021UniFuseUF} & 0.4816 & 94.51 \\
  $\rightarrow$3D60 & Ours & \textbf{0.4637} & \textbf{95.02} \\
  \hline
  Matterport3D & UniFuse-Fusion~\cite{Jiang2021UniFuseUF} & 0.3974 & 96.88 \\
  $\rightarrow$3D60 & Ours & \textbf{0.3629} & \textbf{97.05} \\
  \hline
  Stanford2D3D & OmniDepth~\cite{Zioulis2018OmniDepthDD} & 0.9735 & 71.07 \\
  $\rightarrow$3D60 & Ours & \textbf{0.7761} & \textbf{81.47} \\
  \hline
  Matterport3D & OmniDepth~\cite{Zioulis2018OmniDepthDD} & 0.4363 & 95.29 \\
  $\rightarrow$3D60 & Ours & \textbf{0.4250} & \textbf{95.95} \\
  \hline
  \end{tabular}}
\label{table:generalization}
\vspace{-10pt}
\end{table}

% \begin{table}[t]
% \caption{Comparison of computational costs and time consuming for feature up-sampling between Bicubic interpolation and our CIIF.}
% \setlength{\tabcolsep}{8.5mm}
%   \centering
%   \renewcommand{\arraystretch}{1.2}
%   {
%   \begin{tabular}{c | c | c}
%   \hline
%   ~ & CIIF & Bicubic \\
%   \hline
%   GFLOPs & 0.65 & 1.21 \\
%   \hline
%   Time (ms) & 26.5 & 0.5 \\
%   \hline
%   \end{tabular}}
% \label{table:computaionalcosts}
% \end{table}

% \begin{table}[t]
% \caption{Comparison of computational costs among different input sizes and output sizes.}
% \setlength{\tabcolsep}{1.5mm}
% \renewcommand{\arraystretch}{1.1}
%  \centering
%   {
%   \begin{tabular}{c | c | c | c}
%   \hline
%    Depth Backbone & Input Size & Output Size & GFLOPs \\
%   \hline
%   \multirow{4}{*}{OmniDepth~\cite{Zioulis2018OmniDepthDD}} & $128 \times 256$  & \multirow{2}{*}{$512 \times 1024$ }  & 21  \\ 
%   \cline{2-2} \cline{4-4}
%   ~ & $512 \times 1024$ &  & 317 \\
%   \cline{2-3} \cline{4-4}
%   ~ & $256 \times 512$ & \multirow{2}{*}{$1024 \times 2048$ }  & 86 \\
%   \cline{2-2} \cline{4-4}
%   ~ & $1024 \times 2048$ &  & 1267 \\ \hline
%   \multirow{4}{*}{UniFuse-Fusion~\cite{Jiang2021UniFuseUF}} & $128 \times 256$  & \multirow{2}{*}{$512 \times 1024$ }  & 6  \\ 
%   \cline{2-2} \cline{4-4}
%   ~ & $512 \times 1024$ &  & 63 \\
%   \cline{2-3} \cline{4-4}
%   ~ & $256 \times 512$ & \multirow{2}{*}{$1024 \times 2048$ }  & 22 \\
%   \cline{2-2} \cline{4-4}
%   ~ & $1024 \times 2048$ &  & 250 \\
%   \hline
%   \end{tabular}}
% \label{table:total-cost}
% \end{table}

\noindent \textbf{Different ODI SR backbones.} In Tab.~\ref{table:srbackbone}, we compare the performance with different backbones in the ODI SR task. \ie, SwinIR, Omni-SR, and EDSR-baseline. However, we find that by employing deeper SR backbones, the performance of depth estimation drops. We think the reason is that the deeper SR backbones are not easy to train. Specifically, the Stanford2D3D dataset only contains 1000 training samples, which is not friendly for attention-based backbones.
% For example, with SwinIR as backbone, the Abs Rel metric drops 0.0088 compared with using EDSR-baseline. We think the reason is that the deeper SR backbones are not easy to train. Specifically, the Stanford2D3D dataset only contains 1000 training samples, which is not friendly for attention-based backbones. In addition, the deeper attention-based backbones need more time to converge. Instead, our employed depth estimation backbone, \eg, OmniDepth, would converge in advance in the training process.

\noindent \textbf{Generalization ability.} To verify it, we further test on 3D60 dataset with pre-trained depth estimation models, \ie, OmniDepth and UniFuse-Fusion, from Stanford2D3D and Matterport3D datasets in Tab.~\ref{table:generalization}.  With UniFuse-Fusion as the backbone, our weakly-supervised method outperforms the corresponding fully-supervised method, \eg, 3.7\% gain and 8.7\% gain in MAE metric on Stanford2D3D dataset and Matterport3D dataset, respectively. With OmniDepth as the backbone, our method also shows significant improvements, \eg, 10.40\% and 0.66\% in $\delta_3$ metric on Stanford2D3D and Matterport3D datasets, respectively.

% \noindent \textbf{Computational costs and time consumption.} We compare computational costs during the feature up-sampling process based on the same outputs of the UniFuse-Fusion backbone. As shown in Tab.~\ref{table:computaionalcosts}, CIIF has lower computational costs, benefiting from only learning interpolation weights from the LR feature map, instead of learning both interpolation weights and values. Here the computational costs of Bicubic interpolation include the costs of a followed convolutional layer. Then, our CIIF consumes much time. We think the reason is that the CIIF is achieved by our hand-crafted code, which is slower than convolutional and up-sampling operations in the Pytorch library.

% We further report the computational costs with different input sizes and output sizes in Tab.~\ref{table:total-cost}. With OmniDepth as backbone, to achieve the goal of HR depth estimation with $1024 \times 2048$, our method needs 86 GFLOPs. However, with the input of the original resolution, the computational costs increase to 1267 GFLOPs, which is about 15 times of our method. It demonstrates that our method relieves the computational burden with LR ODI as input during inference.
% Then, we compare the overall computaional costs between our weakly-supervised method and the supervised method. It can be seen that our metod has less computational burdens by removing the UISR branch and UKT module during inference.
\section{Conclusion}
In this paper, to the best of our knowledge, we proposed the first work to estimate the HR omnidirectional depth map directly from an LR ODI, when no HR depth GT map is available. We found that the ODIs and paired depth maps share the common scene structural knowledge. Therefore, we introduced an ODI SR task to boost the performance of the HR depth estimation task. Especially, the textural details in the ODI SR task can interfere with the smoothness of depth predictions. In this case, we introduced uncertainty estimation to the ODI SR specially for extracting scene structural knowledge. Upon this, we proposed the CIIF and FD loss in the SSKT module. The CIIF captures the correlations among neighboring pixels and shares the scene structural knowledge between the two tasks, while the FD loss provides extra structural regularization. Experimental results demonstrated that our weakly-supervised method is 1) efficient without extra inference cost, 2) flexible as a plug-and-play approach, and 3) effective in achieving comparable results with the fully-supervised methods.

\noindent \textbf{Limitation and future work:} In this paper, we achieve the high-resolution $360^{\circ}$ depth estimation task via weakly-supervised learning. In this way, we alleviate the difficulty of acquiring high-resolution $360^{\circ}$ depth ground truth. However, acquiring a $360^{\circ}$ depth map is inherently difficult due to stitching and alignment processes. Recently, the depth foundation models of perspective depth estimation have risen by exploiting large-scale labeled and unlabeled perspective images~\cite{yang2024depth}. In the future, we plan to excavate knowledge from such rich perspective images and depth maps, which is an appealing direction to inspire the $360^{\circ}$ depth estimation.

\newpage
\appendix
\begin{abstract}
    We provide more details of the proposed method in the
supplementary material. Specifically, in Sec.~\ref{sec:CIIF} we explain explicitly the proposed cylindrical implicit interpolation function (CIIF) and its sharing strategy.  Sec.~\ref{sec:fdloss} shows more details in the feature distillation (FD) loss. Then, Sec.~\ref{sec:experiment} shows the more experimental results, including datasets and metrics, visual comparisons, ablation studies, and reports on inference time.
\end{abstract}

\subsection{Cylindrical Implicit Interpolation Function}
\label{sec:CIIF}
\subsubsection{Implicit Image Function}

\begin{figure*}[t]
    \centering
    \includegraphics[width=0.7\textwidth]{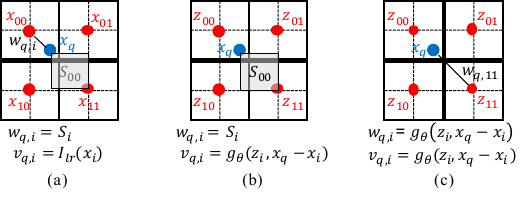}
    \caption{(a) General image interpolation process. (b) LIIF~\cite{Chen2021LearningCI}. (c) JIIF~\cite{tang2021joint}.}
    \label{fig:weight}
\end{figure*}

In general, the image interpolation problem can be defined as follows:
\begin{equation}
    I_{hr}(x_q)=\sum_{i \in N_q}w_{q,i}v_{q,i},
\label{eq-general}
\end{equation}
as shown in Fig.~\ref{fig:weight}(a), where $x_q$ is the coordinate of query pixel $q$ in the HR image domain, $x_i$ is the coordinate of corner pixel $i \in \{00, 01, 10, 11\}$ in the LR image domain and $N_q$ is the set of neighbor pixels for $q$, $w_{q,i}$ is the interpolation weight between $q$ and $i$, and $v_{q,i}$ is the interpolation value. $I_{hr}(x_q)$ is the pixel value of $q$. In traditional image interpolation, the pixel value $I_{lr}(x_i)$ is commonly used as the interpolation value $v_{q,i}$. Due to the nature of perspective images, $N_q$ is usually chosen as the four nearest corner pixels of $q$ in the LR domain (as illustrated in Fig.~\ref{fig:weight}(a), $i \in \{00,01,10,11\}$), and the interpolation weight $w_{q,i}$ is often chosen as the partial area diagonally opposite the corner pixel $i$ (\eg, in Fig.~\ref{fig:weight}(a), interpolation weight $w_{q,00}=S_{00}$), such as Bilinear interpolation.

In implicit image function~\cite{Chen2021LearningCI,tang2021joint}, the image coordinates are normalized into ($-$1, 1) to make it possible to share the coordinate in the LR image domain and the HR image domain. In this way,~\cite{Chen2021LearningCI,tang2021joint} can achieve arbitrary up-sampling factors by querying the pixels in the HR image domain according to corresponding coordinates. For example in Fig.~\ref{fig:weight}(a), we can follow the coordinate $(-0.5,-0.5)$ to query the pixel $x_{00}$ in the LR image domain and follow the coordinate $(-0.125,-0.125)$ to query the pixel $x_q$ in the HR image domain.

\begin{figure}[h]
    \centering
    \includegraphics[width=\linewidth]{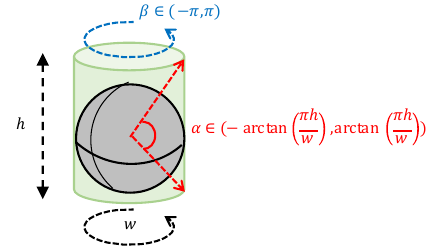}
    \caption{The illustration of cylinder angle coordinate system for ERP images.}
    \label{fig:cylinder}
\end{figure}

% We start from introducing the querying process of IIF, \eg, LIIF~\cite{chen2021learning} and JIIF~\cite{tang2021joint}. As shown in Fig.~\ref{fig:interpolation}, we take the process of interpolating an $2\times2$ image into $8\times8$ image as an example (as Fig~\ref{fig:interpolation} shows). First, all previous IIF methods~\cite{Chen2021LearningCI,tang2021joint} normalize the image coordinates into ($-$1, 1) to make it possible to share the coordinate in the (low-resolution) LR image domain (left 2$\times$2) and HR image (right 8$\times$8). In this way, these IIF methods can achieve arbitrary up-sampling factors by querying the pixels in the HR image domain according to corresponding coordinates. For example in Fig.~\ref{fig:interpolation}, we can follow the coordinate $(-0.5,-0.5)$ to query the left-up pixel (the orange point) in the LR image domain and follow the coordinate $(-0.875,-0.125)$ to query the left-up pixel (the blue point) in the HR image domain.

\begin{figure*}[h]
    \centering
    \includegraphics[width=0.8\textwidth]{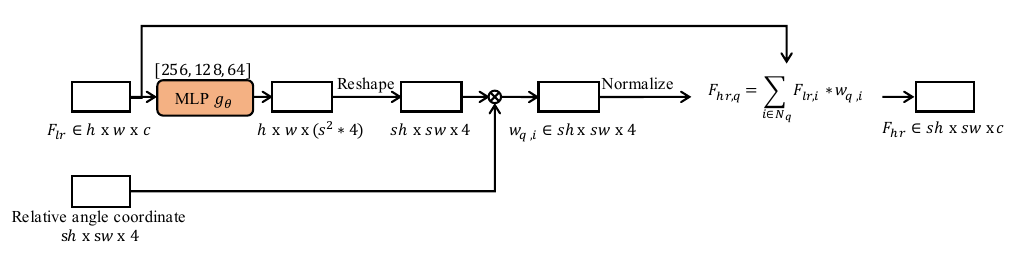}
    \vspace{-5pt}
    \caption{The illustration of the up-sampling process from LR feature to HR feature with our CIIF.}
    \label{fig:up-sample}
    \vspace{-10pt}
\end{figure*}

% \begin{figure}[t]
%     \centering
%     \includegraphics[width=0.8\textwidth]{Figure/supp/interpolation.pdf}
%     \caption{An example for the image interpolation from $2\times2$ to $8\times8$.}
%     \label{fig:interpolation}
% \end{figure}

Furthermore, as shown in Fig.~\ref{fig:weight}(b), in implicit image functions, instead of directly using the pixel value (\eg, RGB value and depth value), the interpolation value $v_{q,i}$ is often generated through a multi-layer perceptron (MLP) $g_{\theta}$ which takes the latent code $z_i$ from an image feature extractor and the relative coordinate $x_q - x_i$ as input, formulated as:

\begin{equation}
    v_{q,i}=g_{\theta}(z_i,x_q-x_i).
    \label{eq:liif}
\end{equation}

% \begin{equation}
%     z^*_{q}=\sum_{i \in N_q}w_{q,i}z_{i}, \ \ z=f_{\theta}(I_{lr}),
%     \label{eq2}
% \end{equation}
% In the LIIF, the interpolation value $v_{q,i}=f_{\phi}([z_i,x_q-x_i])$ while in the
In JIIF~\cite{tang2021joint}, as shown in Fig.~\ref{fig:weight}(c), instead of utilizing the partial area, the interpolation weight $w_{q,i}$ is predicted together with the interpolation value $v_{q,i}$ through an MLP, which views the interpolation weight and the interpolation value as the edge and the node in a graph model, and can be written as:

\begin{equation}
    w_{q,i}, v_{q,i}=g_{\theta}(z_i,x_q-x_i).
    \label{eq:jiif}
\end{equation}

% $v_{q,i}=f_{\phi}([z_i^{hr},z_i^{lr}, x_q-x_i])$ where $z_i^{hr}$ and $z_i^{lr}$ are two latent code maps extracted from the HR guide image and the LR depth. For another parameter weight $w_{q,i}$, LIIF (shown in the Fig~\ref{fig:lweight}) follows bilinear interpolation to set $w_{q,i}=\frac{S_i}{S}$, where $S_i$ is the partial area diagonally opposite to the corner pixel $i$, $S=\sum_{i \in N_q}S_i$ is the total area used as the normalization factor. For JIIF~\cite{Tang2021JointII} (shown in the Fig~\ref{fig:jweight}), the weight $w_{q,i}$ is learned by another MLP to view the interpolation at each pixel as a graph problem.

%$\eta$ with the latent code map of LR depth and the relative latent code as the input and then normalized via the softmax function, so the formulation can be written as  $w_{q,i}=softmax(f_{\eta}(\psi_i,\psi_q-\psi_i))$. Finally, the HR resolution latent code map $z^*$ is decoded by a decoder network.

\subsubsection{Cylindrical Representation}
The aforementioned implicit image functions are designed for perspective images based on Cartesian coordinate system, which are not applicable for omnidirectional images. This is because omnidirectional images are from a sphere, and the most popular equirectangular projection (ERP) is unfold from the sphere based on Cylindrical coordinate system. Using Cartesian coordinate system makes the left side and right side of an ERP image not contiguous. To better represent the spatial correlation of pixels in ERP images, we reframe ERP images with cylindrical representation, as shown in Fig.~\ref{fig:cylinder}. Note that the range of the cylindrical angle coordinate system ($\alpha, \beta$) is:

\begin{equation}
\begin{array}{c}
\alpha\in(-\arctan(\frac{\pi h}{w}), \ \arctan(\frac{\pi h}{w})), \\
\\
\beta\in(-\pi,\ \pi). 
\end{array}
    \label{eq-cylinder}
\end{equation}

In this case, our CIIF can be formulated as $g_{\theta}(z_i,(\alpha_i-\alpha_q,\beta_i-\beta_q))$. As for the output content, we will discuss it in Sec.~\ref{sec:sharing} according to our specific framework and demand.

% \section{Uncertainty-aware Knowledge Transfer (UKT)}

% We propose the uncertainty-aware knowledge transfer (UKT) module to transfer the structural knowledge from the UISR branch to the DESR branch. The proposed UKT module consists of two key components. The first is the sharing strategy that shares the parameters of CIIF as a direct way for transfer, and the second is a feature distillation loss that matches the correlation between the image modality and the depth modality.

\subsubsection{Sharing Strategy}
\label{sec:sharing}

In the above discussion, we have reframed CIIF in cylindrical representation. The aim of CIIF is to transfer structural knowledge from the ODI SR task to the HR depth estimation task. The ODIs and their corresponding depth maps are two different representations for the same scene, which implys that both modals contain the common structural knowledge. Therefore, we directly choose to share parameters of CIIF between the ODI SR task and HR depth estimation task. Note that the parameters of CIIF are all included in an MLP.

As for the output form of CIIF, we find that the predicted interpolation value $v_{q,i}$ in Eq.~\ref{eq:liif} and Eq.~\ref{eq:jiif} can not be shared between different modals. Instead, the relative correlation among neighboring pixels can effectively represent the common structural knowledge of the same scene, depicted as the interpolation weight $w_{q,i}$. In this case, we only output the interpolation weights $w_{q,i}$ from an MLP, formulated as follows:

\begin{equation}
    w_{q,i} = g_{\theta}(z_i,(\alpha_i-\alpha_q,\beta_i-\beta_q)).
\end{equation}

\subsubsection{Feature Up-sampling Process}

As shown in Fig.~\ref{fig:up-sample}, given an LR feature map $F_{lr} \in h \times w \times c$, we aim to obtain an HR feature map $F_{hr} \in sh \times sw \times c$, where $s$ is the up-sampling factor. An MLP with hidden dimension $[256, 128, 64]$ is employed to output interpolation weights $w_{q,i} \in h \times w \times (4*s^2)$. Then, interpolation weights $w_{q,i}$ is reshaped to $sh \times sw \times 4$. We add in the cylindrical relative angle coordinate with the form $\rm{cos}(\alpha_i-\alpha_q)*\rm{cos}(\beta_i-\beta_q)$. We normalize $w_{q,i}$ in the third dimension by $w_{q,i}=\frac{w_{q,i}}{\sum_{i\in4}w_{q,i}}$, and thus each query pixel $q$ in the HR domain has four normalized interpolation weights associated with four corner pixels. Then, $F_{hr}$ is obtained with $w_{q,i}$ and $F_{lr}$ based on Eq.~\ref{eq-general}, where $F_{lr}$ is the interpolation values.

% \begin{figure}[t!]
%     \centering
%     \includegraphics[width=0.8\linewidth]{Figure/FD.pdf}
%     \centering
%     \caption{An illustration of feature distillation (FD) loss.}
%     \label{fig:fdloss}
% \end{figure}

\begin{table}[t!]
\renewcommand{\arraystretch}{1.5}
\setlength{\tabcolsep}{1.2mm}
\caption{The number of samples in training set, validation set, and testing set in three datasets.}
  \centering
  {
  \begin{tabular}{c | c | c | c}
  \hline
 ~ & \multicolumn{3}{c}{Dataset} \\
  \cline{2-4}
  ~ & Stanford2D3D~\cite{Armeni2017Joint2D} & Matterport3D~\cite{chang2017matterport3d}& 3D60~\cite{Zioulis2018OmniDepthDD} \\
  \hline
  Training & 1000 & 7829 & -- \\
  \hline
  Validation & 40 & 947 & -- \\
  \hline
  Testing & 373 & 2014 & 1298 \\
  \hline
  \end{tabular}}
\label{table:datasets}
\end{table}

\subsection{FD Loss}
\label{sec:fdloss}

\begin{figure*}[h]
    \centering
    \includegraphics[width=0.8\textwidth]{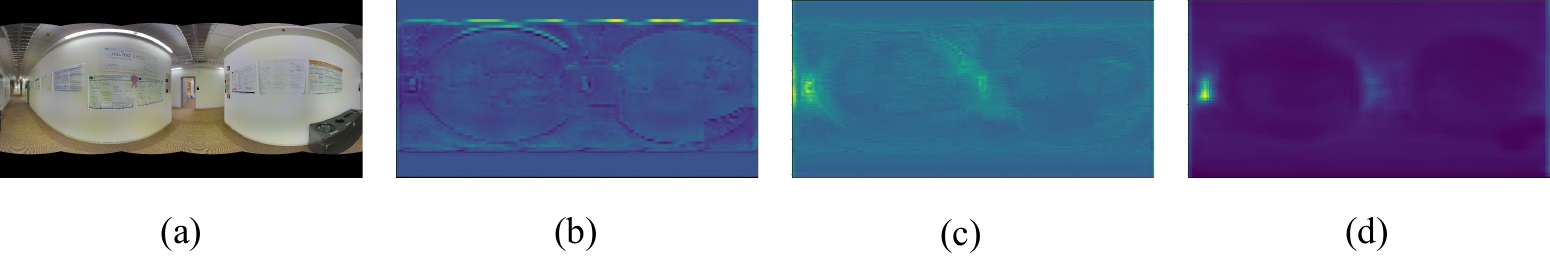}
    \vspace{-10pt}
    \caption{Visualization of the features employed in the FD loss. (a) HR ODI. (b) HR RGB feature from the ODI SR task. (c) HR transformed feature. (d) HR feature from the HR depth estimation task.}
    \label{fig:supp-fdloss}
    \vspace{-15pt}
\end{figure*}

% The overall process of FD loss in shown in Fig.~\ref{fig:fdloss}.
We visualize the HR ODI, the HR RGB feature from the ODI SR task, the HR transformed feature and the HR feature from the HR depth estimation task in Fig.~\ref{fig:supp-fdloss}, respectively. We can find that before transformation, the HR RGB feature from the ODI SR task contains much texture information, which interferes with the smoothness of the HR depth estimation. Instead, the HR transformed feature is more similar with the HR feature from the HR depth estimation task, which is more appropriate for matching the similarity between different tasks. 
% The overall training process of our framework can be seen in Algorithm~\ref{alg}. 

% \begin{table}[t]
% \caption{Ablation studies on FD loss.}
% \setlength{\tabcolsep}{2mm}
% \renewcommand{\arraystretch}{1.2}
%  \centering
%   {
%   \begin{tabular}{c | c | c c}
%   \hline
%    & Transform & Abs Rel $\downarrow$ & $\delta_3 \uparrow$\\
%   \hline
%   \multirow{2}{*}{Feature affinity~\cite{Wang2020DualSL}} &  & 0.1645  & 96.93  \\ 
%   ~ & \checkmark & 0.1630 & \textbf{97.42} \\ \hline
%    \multirow{2}{*}{Cosine similarity} & & 0.1723 & 97.34 \\
%    ~ & \checkmark & \textbf{0.1585} & 97.37 \\
%   \hline
%   \end{tabular}}
% \label{table:ablation-equi}
% \end{table}

\subsection{Experiment}
\label{sec:experiment}

\subsubsection{Dataset and Metrics}

\noindent \textbf{Dataset.} We follow the split files of UniFuse\footnote{\url{https://github.com/alibaba/UniFuse-Unidirectional-Fusion/tree/main/UniFuse/datasets}} to split the three datasets respectively, as depicted in Tab.~\ref{table:datasets}. As the 3D60 dataset is only used to verify the generalization ability, the splits of its training and validation subsets are omitted.

\noindent \textbf{Metrics.} We evaluate our method with standard metrics introduced in~\cite{Jiang2021UniFuseUF}. We only calculate the observed pixels $p \in \mathrm{obs}$, which are valid in the HR depth  GT maps $D^{\mathrm{GT}}_{\mathrm{HR}}$. Notably, during training, our weakly-supervised method only utilizes LR depth maps $D^{\mathrm{GT}}_{\mathrm{LR}}$ as the supervision, while HR depth GT maps $D^{\mathrm{GT}}_{\mathrm{HR}}$ is only utilized for evaluation. Given $D^{\mathrm{GT}}_{\mathrm{HR}}$ and HR depth prediction $D_{\mathrm{HR}}$, the metrics include:

\begin{itemize}
    \item Mean Absolute Error (MAE) measures the absolute difference between the depth prediction and ground truth. It is to show how close between the prediction and ground truth.
    \begin{equation}
    \frac{1}{\mathrm{N}_{\mathrm{obs}}}\sum_{p \in \mathrm{obs}}||D_{\mathrm{HR}}(p) - D^{\mathrm{GT}}_{\mathrm{HR}}(p)||.
    \end{equation}
    
    \item Absolute Relative Error (Abs Rel) measures the absolute difference between the depth prediction and ground truth, scaled by the ground truth. It is to express how large the error is compared with the ground truth.
    \begin{equation}
    \frac{1}{\mathrm{N}_{\mathrm{obs}}}\sum_{p \in \mathrm{obs}}\frac{||D_{\mathrm{HR}}(p) - D^{\mathrm{GT}}_{\mathrm{HR}}(p)||}{D^{\mathrm{GT}}_{\mathrm{HR}}(p)}.
    \end{equation}
    
    \item Root Mean Square Error in log space (RMSE log) measures the square root of the squared difference. The squared difference can emphasize the scenes with large depth variances. In addition, utilizing the log form can emphasize scale differences instead of absolute distance differences.
    \begin{equation}
    \sqrt{\frac{1}{\mathrm{N}_{\mathrm{obs}}}\sum_{p \in \mathrm{obs}}||\log_{10} D_{\mathrm{HR}}(p) - \log_{10} D^{\mathrm{GT}}_{\mathrm{HR}}(p)||^2}.
    \end{equation}

    \item $\delta_i$, the fraction of pixels where the relative error between the depth prediction $D_{\mathrm{HR}}$ and depth GT map $D^{\mathrm{GT}}_{\mathrm{HR}}$ is less than the threshold $1.25^i$, $i \in \{1,2,3\}$. It is to evaluate the accuracy ratios rather than absolute or squared differences.
    \begin{equation}
    max\{\frac{D_{\mathrm{HR}}(p)}{D^{\mathrm{GT}}_{\mathrm{HR}}(p)}, \frac{D^{\mathrm{GT}}_{\mathrm{HR}}(p)}{D_{\mathrm{HR}}(p)}\} < i,
    \end{equation}
\end{itemize}

\noindent where $\mathrm{N}_{\mathrm{obs}}$ represents the total number of the observed pixels in $D^{\mathrm{GT}}_{\mathrm{HR}}$.

\begin{table}[t]
\caption{Ablation studies on Berhu loss.}
\setlength{\tabcolsep}{5.5mm}
\renewcommand{\arraystretch}{1.2}
 \centering
  {
  \begin{tabular}{c | c c}
  \hline
    & Abs Rel $\downarrow$ & $\delta_3 \uparrow$\\
  \hline
  L1 loss  &  0.1609 & 97.18 \\
  Berhu loss~\cite{laina2016deeper}  & \textbf{0.1585} & \textbf{97.37} \\
  \hline
  \end{tabular}}
\label{table:ablation-berhu}
\end{table}

\begin{table}[t]
\caption{Ablation studies on the effect of spherical convolutions.}
\setlength{\tabcolsep}{3.5mm}
\renewcommand{\arraystretch}{1.2}
 \centering
  {
  \begin{tabular}{c | c c c}
  \hline
  EquiConv & MAE $\downarrow$ & Abs Rel $\downarrow$ & $\delta_1 \uparrow$\\
  \hline
  \checkmark  & 0.3144 &  0.1589 & \textbf{79.16} \\
   & \textbf{0.3135} & \textbf{0.1585} & 78.06 \\
  \hline
  \end{tabular}}
\label{table:ablation-equi}
\end{table}

% \begin{table}[!t]
% \caption{Comparison of estimating uncertainty (Ours) with estimating the gradient for the ODI SR task.}
% \setlength{\tabcolsep}{3.0mm}
%   \centering
%   \renewcommand{\arraystretch}{1.2}
%   {
%   \begin{tabular}{c | c c c}
%   \hline
%    & MAE $\downarrow$ & Abs Rel $\downarrow$ & $\delta_1 \uparrow$ \\
%   \hline
%   Gradient & 0.3141 & 0.1635 & \textbf{78.30} \\
%  Uncertainty (Ours) & \textbf{0.3135} & \textbf{0.1585} & 78.06 \\
%   \hline
%   \end{tabular}}
% \label{table:ablation-uncertainty}
% \end{table}

\begin{table}[t]
\caption{Investigation of applying CIIF to the fully-supervised methods.}
\setlength{\tabcolsep}{4.2mm}
\renewcommand{\arraystretch}{1.2}
 \centering
  {
  \begin{tabular}{c | c c c}
  \hline
   & MAE $\downarrow$ & Abs Rel $\downarrow$ & $\delta_1 \uparrow$\\
  \hline
   Conv & 0.3124 &  0.1582 & \textbf{80.16} \\
  CIIF & \textbf{0.3049} & \textbf{0.1577} & 79.66 \\
  \hline
  \end{tabular}}
\label{table:ablation-ciif}
\end{table}

\begin{table}[t]
\caption{Ablation studies on FD loss about different calculations of feature similarity and whether applying transform.}
\setlength{\tabcolsep}{3mm}
\renewcommand{\arraystretch}{1.2}
 \centering
  {
  \begin{tabular}{c | c | c c}
  \hline
   & Transform & Abs Rel $\downarrow$ & $\delta_3 \uparrow$\\
  \hline
  \multirow{2}{*}{Feature affinity~\cite{Wang2020DualSL}} &  & 0.1645  & 96.93  \\ 
  ~ & \checkmark & 0.1630 & \textbf{97.42} \\ \hline
   \multirow{2}{*}{Cosine similarity} & & 0.1723 & 97.34 \\
   ~ & \checkmark & \textbf{0.1585} & 97.37 \\
  \hline
  \end{tabular}}
\label{table:ablation-fd}
\end{table}

\begin{table}[t]
\caption{Discussion about the effect of employing SwinIR, which is a deeper SR backbone.}
\setlength{\tabcolsep}{4mm}
\renewcommand{\arraystretch}{1.2}
 \centering
  {
  \begin{tabular}{c | c c c}
  \hline
  Setting & Abs Rel $\downarrow$ & RMSE log $\downarrow$ & $\delta_3 \uparrow$\\
  \hline
  3  & 0.3414 & 0.2630  & 75.25 \\
   5 & 0.4717 & 0.2653 & 77.73 \\
  \hline
  \end{tabular}}
\label{table:ablation-setting}
\end{table}

\begin{table}[h]
\caption{Comparison of computational costs and time consuming for feature up-sampling between Bicubic interpolation and our CIIF.}
\setlength{\tabcolsep}{8.5mm}
  \centering
  \renewcommand{\arraystretch}{1.2}
  {
  \begin{tabular}{c | c | c}
  \hline
  ~ & CIIF & Bicubic \\
  \hline
  GFLOPs & 0.65 & 1.21 \\
  \hline
  Time (ms) & 26.5 & 0.5 \\
  \hline
  \end{tabular}}
\label{table:computaionalcosts}
\end{table}

\begin{table}[!t]
\caption{Comparison of computational costs among different input sizes and output sizes.}
\setlength{\tabcolsep}{1.5mm}
\renewcommand{\arraystretch}{1.1}
 \centering
  {
  \begin{tabular}{c | c | c | c}
  \hline
   Depth Backbone & Input Size & Output Size & GFLOPs \\
  \hline
  \multirow{4}{*}{OmniDepth~\cite{Zioulis2018OmniDepthDD}} & $128 \times 256$  & \multirow{2}{*}{$512 \times 1024$ }  & 21  \\ 
  \cline{2-2} \cline{4-4}
  ~ & $512 \times 1024$ &  & 317 \\
  \cline{2-3} \cline{4-4}
  ~ & $256 \times 512$ & \multirow{2}{*}{$1024 \times 2048$ }  & 86 \\
  \cline{2-2} \cline{4-4}
  ~ & $1024 \times 2048$ &  & 1267 \\ \hline
  \multirow{4}{*}{UniFuse-Fusion~\cite{Jiang2021UniFuseUF}} & $128 \times 256$  & \multirow{2}{*}{$512 \times 1024$ }  & 6  \\ 
  \cline{2-2} \cline{4-4}
  ~ & $512 \times 1024$ &  & 63 \\
  \cline{2-3} \cline{4-4}
  ~ & $256 \times 512$ & \multirow{2}{*}{$1024 \times 2048$ }  & 22 \\
  \cline{2-2} \cline{4-4}
  ~ & $1024 \times 2048$ &  & 250 \\
  \hline
  \end{tabular}}
\label{table:total-cost}
\end{table}

\begin{figure}[!t]
    \centering
    \includegraphics[width=0.96\linewidth]{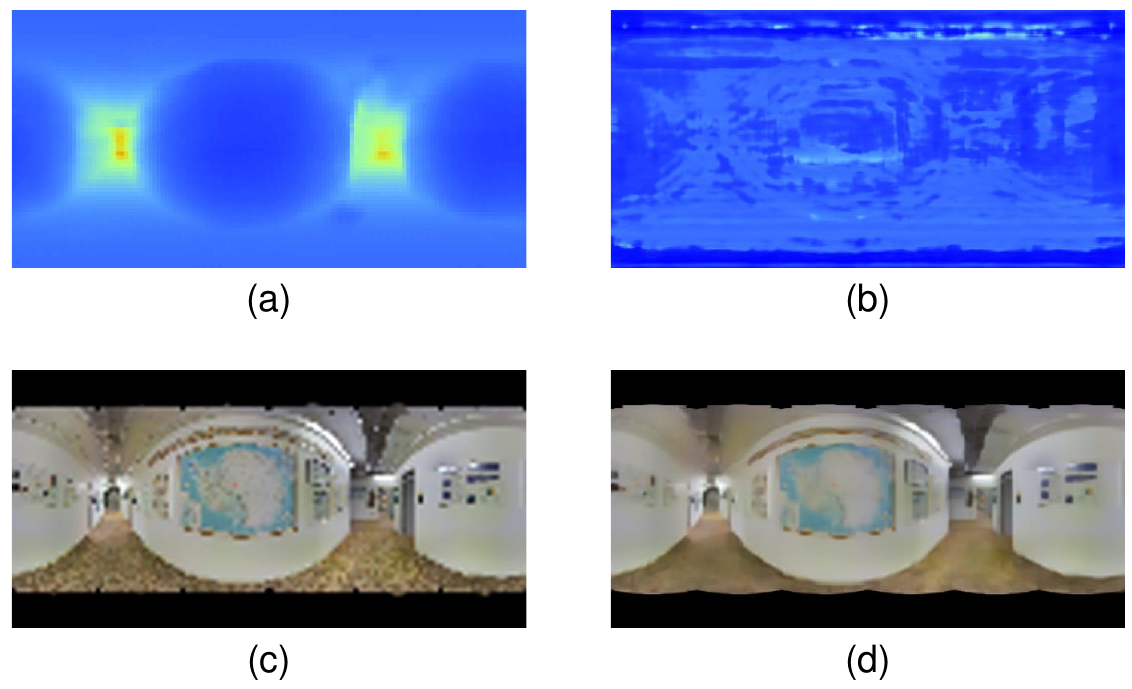}
    \caption{Illustration of the ``noise problem''. (a) The predicted HR depth map in Setting 1; (b) the predicted HR depth map in Setting 3; (c) the LR input in Setting 1; (d) the super-resolved ODI in Setting 3.}
    \label{fig:supp-noise}
\end{figure}

\subsubsection{Ablation Studies}

\noindent \textbf{Berhu loss.} In Tab.~\ref{table:ablation-berhu}, it can be seen that employing Berhu loss~\cite{laina2016deeper} has an improvement in both metrics, such as 0.0024 in Abs Rel metric, which can benefit the depth estimation performance.

\noindent \textbf{Spherical convolutions.} Recently, some efforts have been made to adjust the shape~\cite{Zioulis2018OmniDepthDD} and sampling location~\cite{Tateno2018DistortionAwareCF} of convolution filters to address the distortion problem in ERP type ODI. To verify if the spherical convolution benefits our framework, we choose EquiConvs, which is characterized by their irregular convolutional kernels to adapt to the varying distortions across different latitudes. However, as introduced in BiFuse~\cite{Wang2020BiFuseM3}, the effectiveness of EquiConvs is uncertain, especially in deeper networks. As shown in Table~\ref{table:ablation-equi}, the results show that the influence caused by EquiConvs is minor (0.0009 drop in MAE, and 1.10\% improvement in $\delta_1$). Therefore, it is reasonable that our design in knowledge transfer is more important than applying the EquiConvs.

\noindent \textbf{FD loss.} We further investigate the choice of different calculations of feature similarity in Tab.~\ref{table:ablation-fd}. Without transform, the feature affinity loss~\cite{Wang2020DualSL} even performs worse than cosine-similarity. After adding the transform process, the feature affinity loss improves in both metrics, such as 0.49\% in $\delta_3$ metric. The results show that the transform process is more important than the selection of the calculation of feature similarity.

\subsubsection{Discussion}

% \noindent \textbf{Structural representation.} To investigate the form of structural representation, we replace uncertainty estimation with gradient estimation~\cite{ma2020structure}. From Tab.~\ref{table:ablation-uncertainty}, we can see uncertainty estimation performs better than gradient estimation. 
% Compared with gradient estimation that can only match fixed edges of ODIs and have a certain data distribution~\cite{ning2021uncertainty}, uncertainty estimation is derived from data itself and thus more flexible to fit with the data distribution of depth modal for structural knowledge transfer.

\noindent \textbf{Adding CIIF to current methods.} In the main paper, the fully-supervised methods are learnable up-sampling convolutional layers, such as combining Bicubic up-sampling and convolution layers, which is used in UniFuse. In Tab.~\ref{table:ablation-ciif}, we try to replace the learnable up-sampling layers with our CIIF. We find that the performance of the fully-supervised can be improved significantly, such as 0.0075 in MAE metric. It demonstrates the effectiveness of the CIIF.

\noindent \textbf{Different ODI SR backbones.} In Setting 3 and 5, the choice of deeper SR backbones might provide more accurate SR ODI for the followed depth estimation network. However, In Tab.~\ref{table:ablation-setting}, we find that the depth estimation results in Setting 3 and 5 are still bad. To better illustrate the noise problem, we provide the visualization results in Fig.~\ref{fig:supp-noise}. We observe that the predicted HR depth map from the LR ODI in Setting 1 is obviously better than the predicted result from the SR ODI in Setting 3. In contrast, our framework do not utilize the SR image, but utilize the SR features for knowledge distillation.

\noindent \textbf{Computational costs and inference time.} We further report the computational costs with different input sizes and output sizes in Tab.~\ref{table:total-cost}. With OmniDepth as backbone, to achieve the goal of HR depth estimation with $1024 \times 2048$, our method needs 86 GFLOPs. However, with the input of the original resolution, the computational costs increase to 1267 GFLOPs, which is about 15 times of our method. It demonstrates that our method relieves the computational burden with LR ODI as input during inference.

As shown in Fig.~\ref{fig:supp-cost}, we calculate computational costs with $\times8$ up-sampling factor, EDSR-baseline~\cite{lim2017enhanced} as the feature extractor in the ODI SR task, and UniFuse-Fusion~\cite{Jiang2021UniFuseUF} as the feature extractor in the HR depth estimation task. During training, the feature extractor in the ODI SR task has 9.829 GFLOPs, and the feature extractor in the HR depth estimation task has 0.977 GFLOPs. CIIF which serves for up-sampling feature maps consumes 0.654 GFLOPs. The decoder of each modal is also reported. Once training is done, the ODI SR task can be freely removed, introducing no extra inference cost. Therefore, the overall inference cost is only 1.706 GFLOPs.

\subsubsection{More visual results.} 
\begin{figure*}[!t]
    \centering
\includegraphics[width=0.6\textwidth]{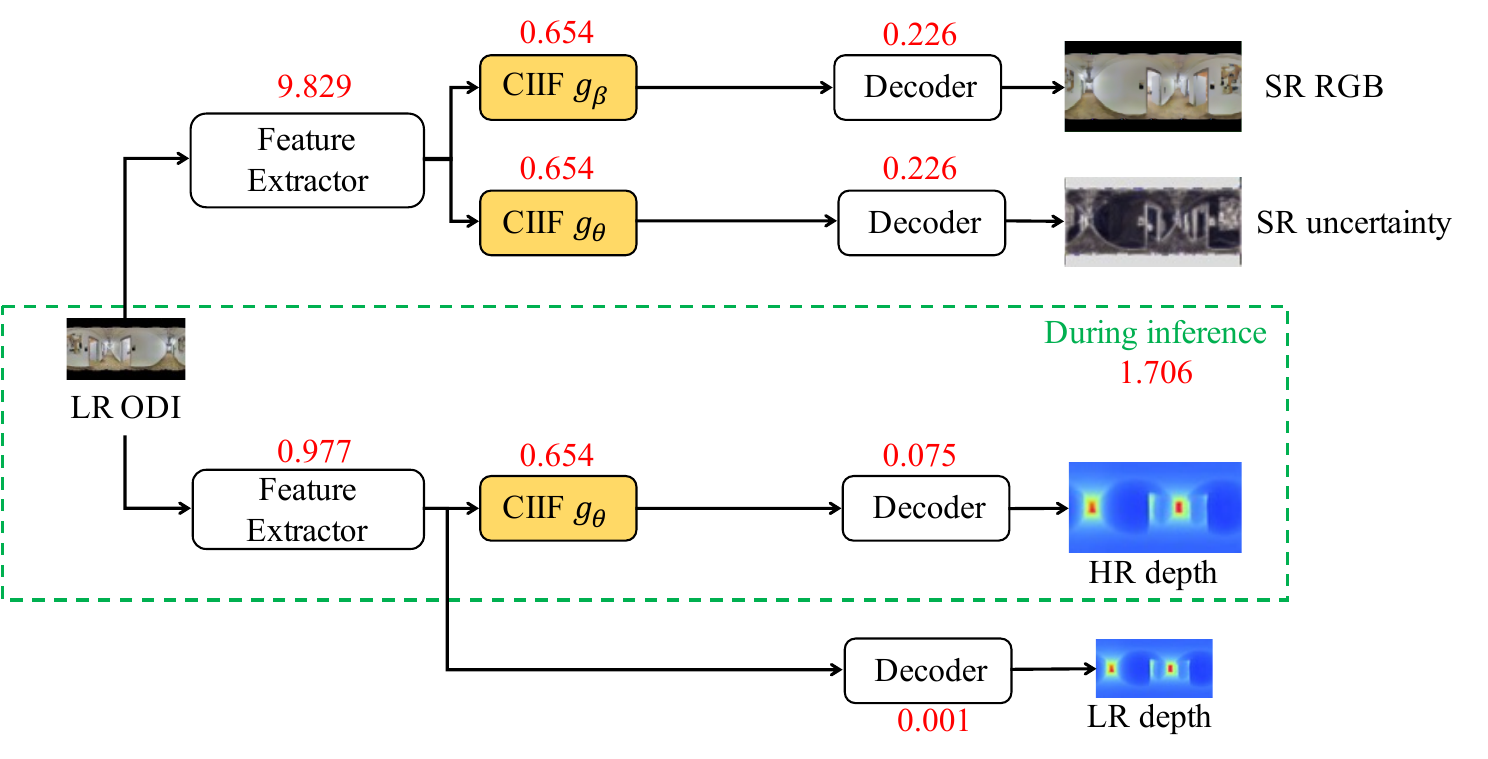}
    \caption{Overview of computational costs of each module.}
    \label{fig:supp-cost}
\end{figure*}

\begin{figure*}[t]
    \centering
    \includegraphics[width=0.8\linewidth]{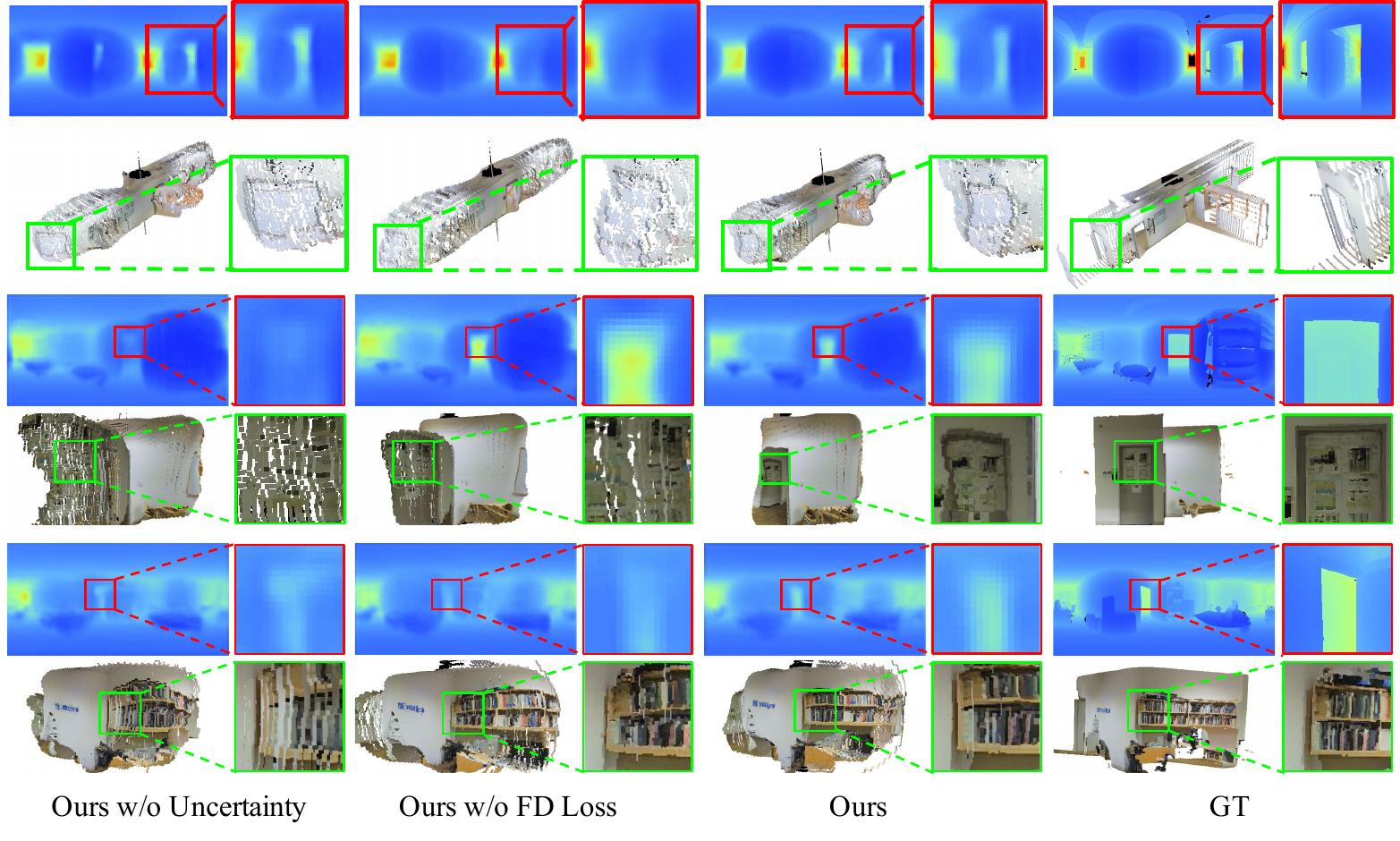}
    \caption{\textbf{Visual comparison of the impacts of uncertainty estimation in the ODI SR task and FD loss} on the overall framework.}
    \label{fig:ablation}
\end{figure*}

In Fig.~\ref{fig:supp-cost}, we provide more visual results on Matterport3D dataset. In addition, in Fig.~\ref{fig:ablation}, we provide visual comparisons about ablation studies.

\begin{figure*}[!t]
    \centering
    \includegraphics[width=0.8\textwidth]{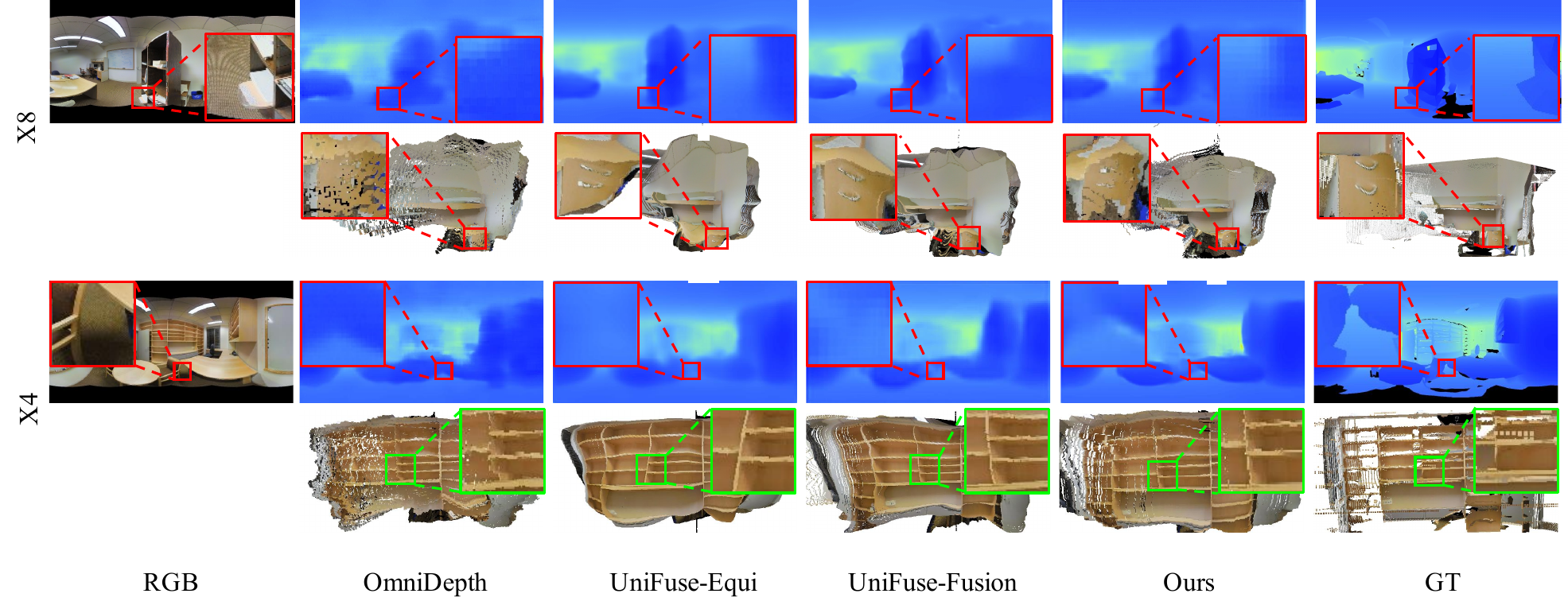}
    \caption{\textbf{Visual comparison of fully-supervised methods and ours} on Stanford2D3D dataset. Best viewed in color.}
    \label{fig:compare_stanford}
\end{figure*} 

% \begin{figure}[!t]
%     \centering
%     \includegraphics[width=0.94\linewidth]{Figure/intro.pdf}
%     \caption{\textbf{Visual comparison of omnidirectional monocular depth estimation results with $\times8$ up-sampling factor on Stanford2D3D dataset}~\cite{Armeni2017Joint2D}. Left: predicted depth maps, Right: reconstructed point clouds. (a) Our method without uncertainty estimation. (b) Our method without FD loss. (c) Our method (All). (d) UniFuse-Fusion~\cite{Jiang2021UniFuseUF} with fully supervision. (e) Ground truth.}
%     \label{fig:intro}
% \end{figure}

%%%%%%%%% REFERENCES
\bibliographystyle{IEEEtran}
\bibliography{IEEEabrv, egbib}

\begin{IEEEbiography}[{\includegraphics[width=1in,height=1.25in,clip,keepaspectratio]{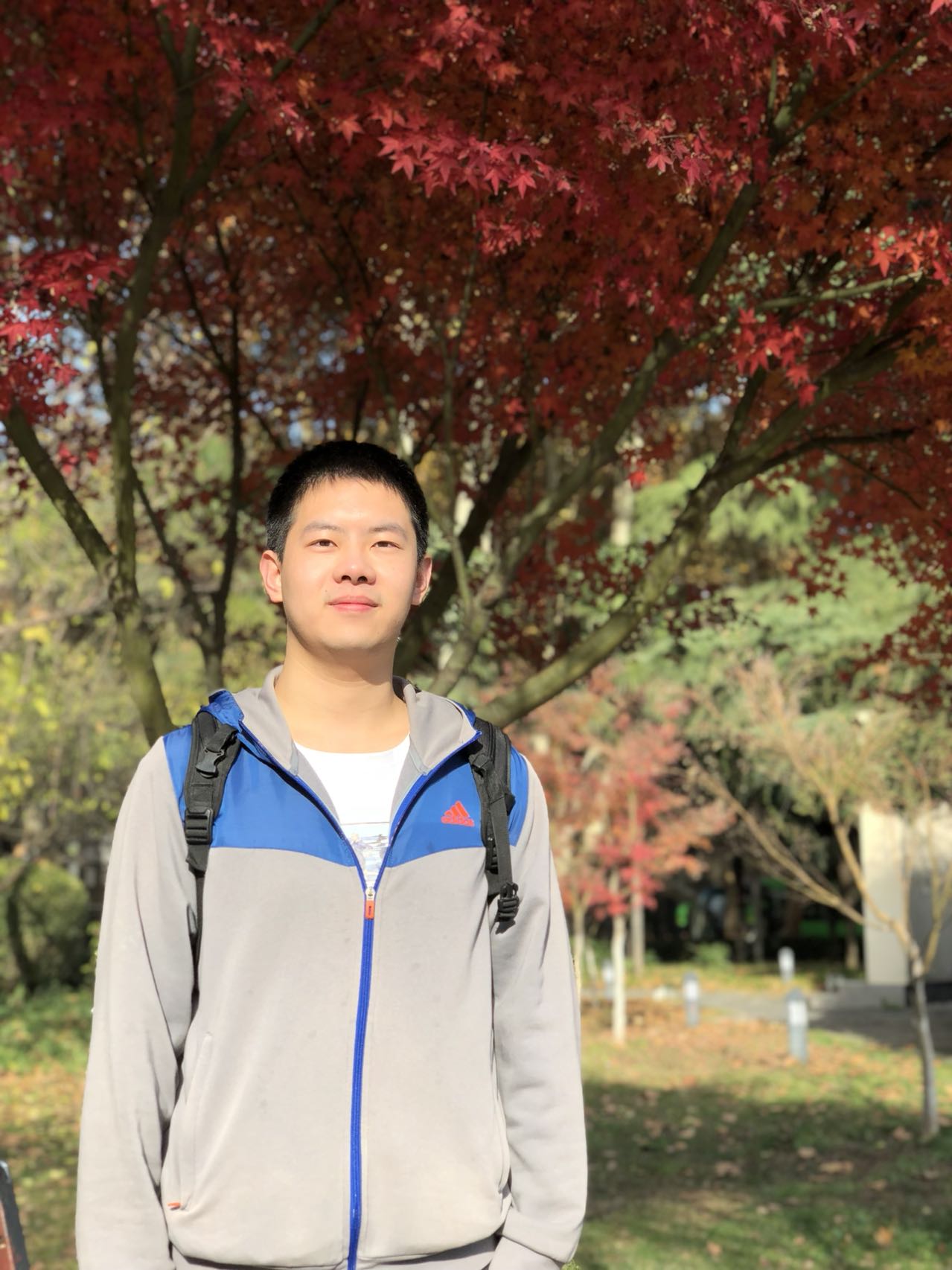}}]{Zidong Cao}
is a Ph.D. student in the Visual Learning and Intelligent Systems Lab,  Artificial Intelligence
Thrust,  Guangzhou Campus, The Hong Kong University of Science and Technology (HKUST). His research interests include 3D vision (depth completion, depth estimation, etc.), DL (self-supervised learning, weakly-supervised learning, etc.), and omnidirectional vision.
\end{IEEEbiography}

\begin{IEEEbiography}[{\includegraphics[width=1in,height=1.25in,clip,keepaspectratio]{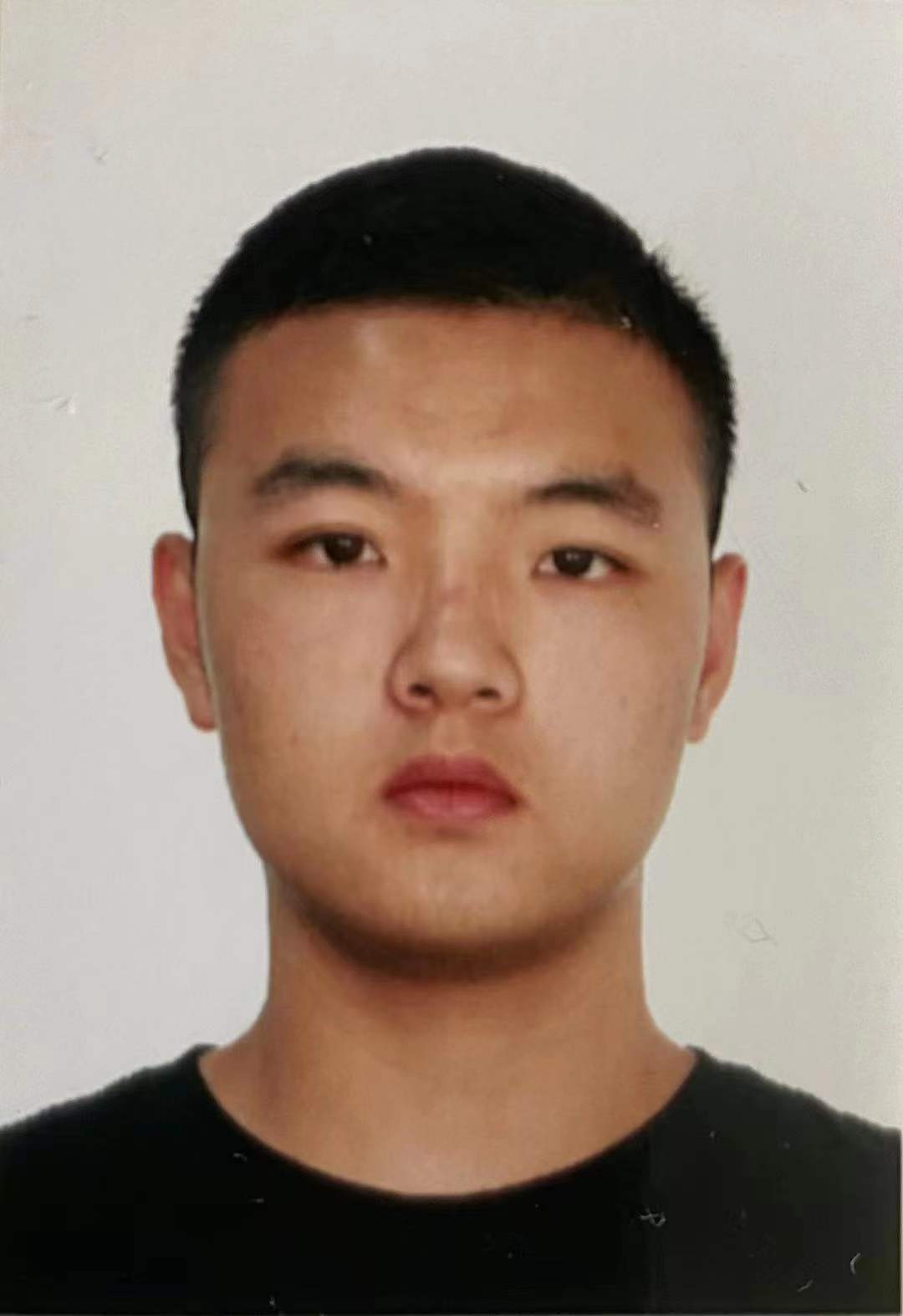}}]{Hao Ai}
is a Ph.D. student in the Visual Learning and Intelligent Systems Lab,  Artificial Intelligence
Thrust,  Guangzhou Campus, The Hong Kong University of Science and Technology (HKUST). His research interests include pattern recognition (image classification, face recognition, etc.), DL (especially uncertainty learning, attention, transfer learning, semi- /self-unsupervised learning), and omnidirectional vision.
\end{IEEEbiography}

% \begin{IEEEbiography}[{\includegraphics[width=1in,height=1.25in,clip,keepaspectratio]{Figure/wanglin.jpg}}]{Lin Wang} (IEEE Member) is an assistant professor in the AI Thrust, HKUST-GZ, HKUST FYTRI, and an affiliate assistant professor in the Dept. of CSE, HKUST. He is the director of the Visual Learning and Intelligent Systems Lab. He did his Postdoc at the Korea Advanced Institute of Science and Technology (KAIST). He got his Ph.D. (with honors) and M.S. from KAIST, where he had research experience in Industrial, Computer, and Mechanical Engineering. His research interests lie in computer and robotic vision, machine learning, intelligent systems, human-AI interaction, AI for science, etc.  
% \end{IEEEbiography}

\begin{IEEEbiography}[{\includegraphics[width=0.8in,height=1.0in,clip, keepaspectratio]{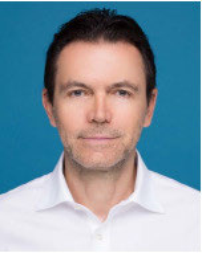}}] 
{Athanasios V. Vasilakos} is with the Center for AI Research (CAIR), University of Agder(UiA), Grimstad, Norway. He served or is serving as an Editor for many technical journals, such as the IEEE TRANSACTIONS ON AI, IEEE TRANSACTIONS ON NETWORK AND SERVICE MANAGEMENT; IEEE TRANSACTIONS ON CLOUD COMPUTING, IEEE TRANSACTIONS ON INFORMATION FORENSICS AND SECURITY, IEEE TRANSACTIONS ON CYBERNETICS; IEEE TRANSACTIONS ON NANOBIOSCIENCE; IEEE TRANSACTIONS ON INFORMATION TECHNOLOGY IN BIOMEDICINE; ACM Transactions on Autonomous and Adaptive Systems; the IEEE JOURNAL ON SELECTED AREAS IN COM-MUNICATIONS. He is WoS highly cited researcher(HC).
\end{IEEEbiography}

\begin{IEEEbiography}[{\includegraphics[width=1in,height=1.25in,clip,keepaspectratio]{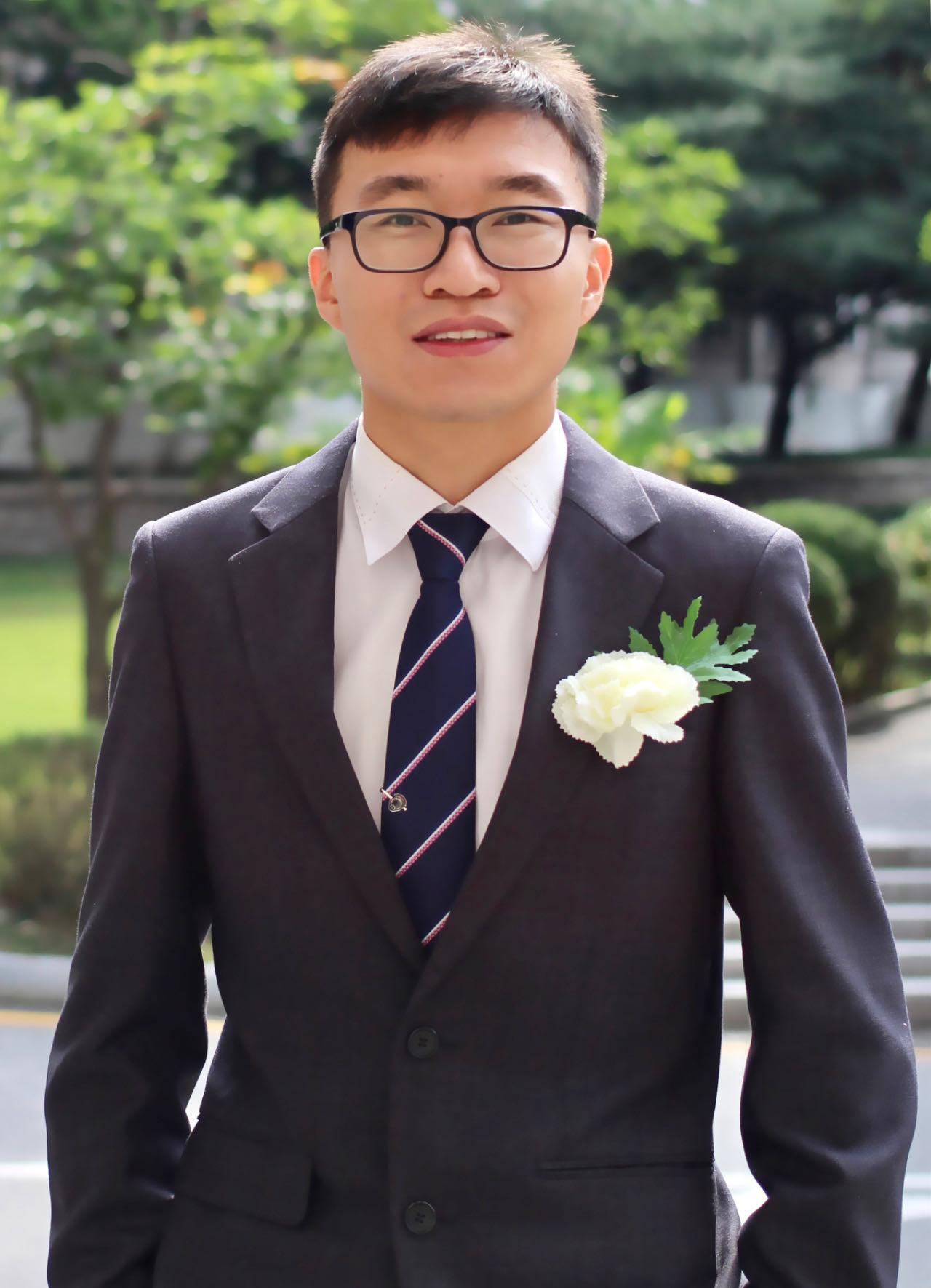}}] 
{Lin Wang} (IEEE Member) is an assistant professor in the AI Thrust, CMA Trust at HKUST, GZ, and an affiliate assistant professor in the Dept. of CSE, HKUST, CWB. He did his Postdoc at the Korea Advanced Institute of Science and Technology (KAIST). He got his Ph.D. (with highest honors) and M.S. from KAIST, Korea. He had rich cross-disciplinary research experience, covering mechanical, industrial, and computer engineering. His research interests lie in computer and robotic vision, machine learning, intelligent systems (XR, vision for HCI), etc. For more about me, please visit https://vlislab22.github.io/vlislab/.
\vspace{-50pt}
\end{IEEEbiography}

\end{document}